\documentclass[]{article}

\usepackage{hyperref}       
\usepackage{url}            
\usepackage{booktabs}       
\usepackage{amsfonts}       

\usepackage{graphicx}
\usepackage{subcaption}

\usepackage{tcolorbox}
\definecolor{darkred}{RGB}{150,0,0}
\definecolor{darkgreen}{RGB}{0,150,0}
\definecolor{darkblue}{RGB}{0,0,150}
\hypersetup{colorlinks=true, linkcolor=red, citecolor=darkgreen, urlcolor=darkblue}

\usepackage{tikz}
\usepackage{amssymb}
\usepackage{amsmath}
\usepackage{amsmath,amsfonts,amssymb,amsthm}
\usepackage{mathtools}
\usepackage{commath}
\usepackage{flexisym}
\usepackage[utf8]{inputenc}
\usepackage{csquotes}
\usepackage[english]{babel}

\usepackage{color}
\usepackage{bbm}

\usepackage{amsmath}
\DeclareMathOperator*{\argmax}{arg\,max}

\newtheorem{lemma}{Lemma}
\newtheorem{corollary}{Corollary}
\newtheorem{theorem}{Theorem}
\newtheorem{myassum}{Assumption}
\theoremstyle{definition}

\theoremstyle{definition}
\newtheorem{definition}{Definition}
\newtheorem{proposition}{Proposition}

\usepackage[shortlabels]{enumitem}
\usepackage{lipsum}

\usepackage{enumitem}
\usepackage{xcolor}
\usepackage[linesnumbered,ruled,vlined]{algorithm2e}

\SetCommentSty{mycommfont}

\SetKwInput{KwInput}{Input}                
\SetKwInput{KwOutput}{Output} 

\usepackage[margin=0.75in]{geometry}

\title{Safe Reinforcement Learning with Linear Function Approximation}

\newcommand{\SUCB}{SLUCB-QVI}
\newcommand{\RSUCB}{RSLUCB-QVI}

\newcommand{\la}{\lambda}

\newcommand{\nn}{\nonumber}

\newcommand{\bal}{\begin{align}}
\newcommand{\eal}{\end{align}}

\DeclarePairedDelimiterX{\inp}[2]{\langle}{\rangle}{#1, #2}

\newcommand{\phib}{\boldsymbol{\phi}}

\newcommand{\A}{\mathbf{A}}

\newcommand{\Vb}{\mathbf{V}}

\newcommand{\x}{\mathbf{x}}

\newcommand{\w}{\mathbf{w}}

\newcommand{\bb}{\mathbf{b}}
\newcommand{\e}{\mathbf{e}}
\newcommand{\y}{\mathbf{y}}

\newcommand{\rb}{\mathbf{r}}

\newcommand{\Sc}{{\mathcal{S}}}

\newcommand{\Vc}{\mathcal{V}}
\newcommand{\Dc}{\mathcal{D}}

\newcommand{\Nc}{\mathcal{N}}

\newcommand{\Cc}{\mathcal{C}}

\newcommand{\Ac}{\mathcal{A}}

\newcommand{\Ec}{\mathcal{E}}

\newcommand{\Oc}{\mathcal{O}}

\newcommand{\beq}{\begin{equation}}
\newcommand{\eeq}{\end{equation}}
\newcommand{\bea}{\begin{align}}
\newcommand{\eea}{\end{align}}

\newcommand{\Otilde}{\tilde\Oc}

\newcommand{\Pb}{\mathbb{P}}
\newcommand{\Ahk}{\Ac_h^k}
\newcommand{\Ah}{\Ac_h}
\newcommand{\Psafe}{\Pi^{\rm safe}}

\newcommand{\mub}{\boldsymbol \mu}

\newcommand{\nub}{\boldsymbol \nu}

\newcommand{\thetab}{\boldsymbol\theta}

\newcommand{\gammab}{\boldsymbol\gamma}

\usepackage{authblk}
\author[1]{Sanae Amani}
\author[2]{Christos Thrampoulidis}
\author[3]{Lin F. Yang}
\affil[1,3]{University of California, Los Angeles}
\affil[2]{University of British Columbia, Vancouver}
{
    \makeatletter
    \renewcommand\AB@affilsepx{, \protect\Affilfont}
    \makeatother
    \affil[1]{samani@ucla.edu}
\affil[2]{cthrampo@ece.ubc.ca}
\affil[3]{linyang@ee.ucla.edu}
}

\begin{document}
\sloppy
\date{}
\maketitle

\begin{abstract}
 Safety in reinforcement learning has become increasingly important in recent years. Yet, existing solutions either fail to strictly avoid choosing unsafe actions, which may lead to catastrophic results in safety-critical systems, or fail to provide regret guarantees for settings where safety constraints need to be learned. In this paper, we address both problems by first modeling safety as an unknown linear cost function of states and actions, which must always fall below a certain threshold. We then present algorithms, termed \SUCB~and \RSUCB, for finite-horizon Markov decision processes (MDPs) with linear function approximation. We show that \SUCB~and \RSUCB, while with \emph{no safety violation}, achieve a $\Otilde\left(\kappa\sqrt{d^3H^3T}\right)$ regret, nearly matching that of state-of-the-art unsafe algorithms, where $H$ is the duration of each episode, $d$ is the dimension of the feature mapping, $\kappa$ is a constant characterizing the safety constraints, and $T$ is the total number of action played. We further present numerical simulations that corroborate our theoretical findings.
\end{abstract}

\section{Introduction}
Reinforcement Learning (RL) is the study of an agent trying to maximize its expected
cumulative reward by interacting with an unknown environment over time \cite{sutton2018reinforcement}. In most classical RL algorithms, agents aim to maximize a long term gain by exploring all possible actions. However, freely exploring all actions may be harmful in many real-world systems where playing even one unsafe action may lead to catastrophic results. Thus, safety in RL has become a serious issue that restricts the applicability of RL algorithms to many real-world systems. For example, in a self-driving car, it is critical to explore those policies that avoid crash and damage to the car, people and property. Switching cost limitations in medical applications \cite{bai2019provably} and legal restrictions in financial managements \cite{abe2010optimizing} are other examples of safety-critical applications. All the aforementioned safety-critical environments introduce the new challenge of balancing the goal of reward maximization with the restriction of playing safe actions.

To address this major concern, the learning algorithm needs to guarantee that it does not violate certain safety constraints. From a bandit optimization point of view, \cite{amani2019linear,pacchiano2020stochastic,Amani_Thrampoulidis_2021,moradipari2019safe} study a linear bandit problem, in which, at each round, a linear cost constraint needs to be satisfied with high probability. For this problem, they propose no-regret algorithms that with high probability never violate the constraints. There has been a surge of research activity to address the issue of safe exploration in RL when the environment is modeled via the more challenging and complex setting of an unknown MDP.
Many of existing algorithms model the safety in RL via Constrained Markov Decision Process (CMDP), that extends the classical MDP to settings with extra constraints on the total expected cost over a horizon. To address the safety requirements in CMDPs, different approaches such as Primal-Dual Policy Optimization \cite{paternain2019constrained,paternain2019safe,stooke2020responsive}, Constrained Policy Optimization \cite{achiam2017constrained,yang2020projection}, and Reward
Constrained Policy Optimization \cite{tessler2018reward} have been proposed. These algorithms come with either no theoretical
guarantees or asymptotic convergence guarantee in the batch offline setting. In another line of work studying CMDP in online settings, \cite{efroni2020exploration,turchetta2020safe,garcelon2020conservative,zheng2020constrained,ding2020provably,qiu2020upper,ding2020natural,xu2020primal,kalagarla2020sample} propose algorithms coming with sub-linear bounds on the number of constraint violation. Additionally, the safety constraint considered in the aforementioned papers is defined by the cumulative expected cost over a horizon falling below a certain threshold. 

In this paper, we propose an upper confidence bound (UCB)- based algorithm -- termed Safe Linear UCB Q/V Iteration (\SUCB) -- with the focus on deterministic policy selection respecting a more restrictive notion of safety requirements that must be satisfied at each time-step an action is played with high probability.  We also present Randomized \SUCB~(\RSUCB), a safe algorithm focusing on randomized policy selection without any constraint violation. For both algorithms, we assume the underlying MDP has linear structure and prove a regret bound that is order-wise comparable to those of its unsafe counter-parts.

Our main technical contributions allowing us to guarantee sub-linear regret bound while the safety constraints are never violated, include: 1) conservatively selecting actions from properly defined subsets of the unknown safe sets; and 2) exploiting careful algorithmic designs to ensure \emph{optimism in the face of safety constraints}, i.e., the value function of our proposed algorithms are greater than the optimal value functions. See Sections \ref{sec:SUCB},\ref{sec:theoreticalguarantees}, and \ref{sec:randomizedSUCB} for details.

\textbf{Notation.} 
We start by introducing a set of notations that
are used throughout the paper. We use lower-case letters for scalars, lower-case bold letters for vectors, and upper-case bold letters for matrices. The Euclidean-norm of $\x$ is denoted by $\norm{\x}_2$. We denote the transpose of any column vector $\x$ by $\x^\top$. For any vectors $\x$ and $\y$, we use $\langle \x,\y\rangle$ to denote their inner product. Let $\A$ be a positive definite $d\times d$ matrix and $\boldsymbol \nu \in\mathbb R^d$. The weighted 2-norm of $\boldsymbol \nu$ with respect to $\A$ is defined by $\norm{\boldsymbol \nu}_\A = \sqrt{\boldsymbol \nu^\top \A \boldsymbol \nu}$. For positive integer $n$, $[n]$ denotes the $\{1,2,\ldots,n\}$. We use $\mathbf{e}_i$ to denote the $i$-th standard basis vector. Finally, we use standard $\Otilde$ notation for big-O notation that ignores logarithmic factors.
\subsection{Problem formulation} \label{sec:formulate}
{\bf Finite-horizon Markov decision process.} We consider a finite-horizon Markov decision process (MDP) denoted by $M=(\Sc,\Ac, H,\Pb, r, c)$, where $\Sc$ is the state set, $\Ac$ is the action set, $H$ is the length of each episode (horizon), $\Pb=\{\Pb_h\}_{h=1}^H$ are the transition probabilities, $r=\{r_h\}_{h=1}^H$ are the reward functions, and  $c=\{c_h\}_{h=1}^H$ are the safety measures. For each time-step $h\in[H]$, $\Pb_h(s'|s,a)$ denotes the probability of transitioning to state $s'$ upon playing action $a$ at state $s$, and $r_h:\Sc\times\Ac\rightarrow [0,1]$ and $c_h:\Sc\times\Ac\rightarrow [0,1]$ are reward and constraint functions. We consider the learning problem where $\Sc$ and $\Ac$ are known, while the transition
probabilities $\Pb_h$, rewards $r_h$ and safety measures $c_h$ are \emph{unknown} to the agent and must be learned
online. The agent interacts with its unknown environment described by $M$ in episodes. In particular, at each episode $k$ and time-step $h\in[H]$, the agent observes the state $s_h^k$, plays an action $a_h^k\in\Ac$, and observes a reward $r_h^k:=r_h(s_h^k,a_h^k)$ and a noise-perturbed safety measure $z_h^k:=c_h(s_h^k,a_h^k)+\epsilon_h^k$, where $\epsilon_h^k$ is a random additive noise.

{\bf Safety Constraint.}
We assume that the underlying system is safety-critical and the learning environment is subject to a side constraint that restricts the choice of actions. At each episode $k$ and time-step $h\in[H]$, when being in state $s_h^k$, the agent must select a \emph{safe} action $a_h^k$ such that 
\begin{align}\label{eq:safetyconstraintdeterministic}
    c_h(s_h^k,a_h^k)\leq \tau
\end{align}
with high probability, where $\tau$ is a known constant. We accordingly define the \emph{unknown} safe action sets as
\begin{align}
 \Ah^{{\rm safe}}(s):=\{a\in\Ac\,:\, c_h(s,a)\leq \tau\},\quad \forall (s,h)\in\Sc\times[H].\nn
\end{align}
Thus, after observing state $s_h^k$ at episode $k$ and time-step $h\in[H]$, the agent's choice of action must belong to $\Ah^{\rm safe}(s_h^k)$ with high probability. As a motivating example, consider a self-driving car. On the one hand, the agent (car) is rewarded for getting from point one to point two as fast as possible. On the other hand, the driving behavior must be constrained to respect traffic safety standards.

{\bf Goal.}
A \emph{safe} deterministic policy is a function $\pi:\Sc\times [H]\rightarrow \Ac$, such that $\pi(s,h)\in \Ah^{{\rm safe}}(s)$ is the \emph{safe} action the policy $\pi$ suggests the agent to play at time-step $h\in[H]$ and state $s\in\Sc$. Thus, we define the set of safe policies by
\begin{align}
    \Psafe:=\left\{\pi:\pi(s,h)\in\Ac_h^{\rm safe}(s),~\forall(s,h)\in \Sc\times[H]\right\}.\nn
\end{align}
For each $h\in[H]$, the cumulative expected reward obtained under a safe policy $\pi\in\Psafe$ during and after time-step $h$, known as the value function $V_h^\pi:\Sc\rightarrow\mathbb{R}$, is defined by
\begin{align}\label{eq:valuefunction}
    V_h^\pi(s):=\mathbb{E}\left[\left.\sum_{h'=h}^H r_{h'}\left(s_{h'},\pi(s_{h'},h')\right)\right\vert s_h=s\right],
\end{align}
where the expectation is over the environment. We also define the state-action value action $Q_h^\pi:\Sc\times \Ah^{\rm safe}(.)\rightarrow\mathbb{R}$ for a safe policy $\pi\in \Psafe$ at time-step $h\in[H]$ by
\begin{align}\label{eq:actionstatevaluefunction}
    Q_h^\pi(s,a):=\mathbb{E}\left[\left.\sum_{h'=h+1}^H r_{h'}\left(s_{h'},\pi(s_{h'},h')\right)\right\vert s_h=s, a_h=a\right].
\end{align}
To simplify the
notation, for any function $f$, we denote $[\mathbb{P}_hf](s,a):=\mathbb{E}_{s^\prime\sim\mathbb{P}_h(.|s,a)}f(s^\prime)$.
Let $\pi_\ast$ be the optimal \emph{safe} policy such that $V^{\pi_\ast}_h(s):=V^{\ast}_h(s) = \sup_{\pi\in\Psafe}V_h^\pi(s)$ for all $(s,h)\in\Sc\times [H]$. Thus, for all $(s,h)\in\Sc\times[H]$ and $a\in\Ah^{{\rm safe}}(s)$, the Bellman equations  for an arbitrary safe policy $\pi\in\Psafe$ and the optimal safe policy are:
\begin{align}
    Q_h^\pi(s,a)=r_h(s,a)+[\mathbb{P}_hV^\pi_{h+1}](s,a),\quad V_h^\pi(s)=Q_h^\pi(s,\pi(s,h)),\label{eq:bellmanforpi}\\
Q_h^\ast(s,a)=r_h(s,a)+[\mathbb{P}_hV^\ast_{h+1}](s,a),\quad
V_h^\ast(s)=\max_{a\in\Ac_h^{\rm safe}(s)}Q_h^\ast(s,a)\label{eq:bellmanforoptimal},
\end{align}
where $V_{H+1}^\pi(s)=V_{H+1}^\ast(s)=0$. Note that in classical RL without safety constraints, the Bellman optimality equation implies that there exists at least one optimal policy that is deterministic (see \cite{bertsekas2000dynamic,szepesvari2010algorithms,sutton2018reinforcement}). When considering solving the Bellman equation for the optimal policy, the presence of safety constraints is equivalent to solving it for an MDP without constraints but with different action sets for each $(s,h)\in\mathcal{S}\times[H]$, i.e., $\mathcal{A}_h^{\text{safe}}(s)$.

Let $K$ be the total number of episodes, $s_1^k$ be the initial state at the beginning of episode $k\in[K]$ and $\pi_k$ be the high probability \emph{safe} policy chosen by the agent during episode $k\in[K]$. Then the \emph{cumulative pseudo-regret} is defined by 
\begin{align}\label{eq:regret}
   R_K:=\sum_{k=1}^K V_1^{\ast}(s_1^k)-V_1^{\pi_k}(s_1^k).
\end{align}

The agent's goal is to keep $R_K$ as small as possible ($R_K/K\rightarrow 0$ as $K$ grows large) \emph{without violating the safety constraint in the process}, i.e., $\pi_k\in\Psafe$ for all $k\in[K]$ with high probability.

{\bf Linear Function Approximation.} We focus on MDPs with linear transition kernels, reward, and cost functions that are encapsulated in the following assumption.

\begin{myassum}[Linear MDP \cite{bradtke1996linear,yang2019sample,jin2020provably}]\label{assum:linearMDP}
$M=(\Sc,\Ac, H,\Pb, r, c)$ is a linear MDP with feature map $\phib:\Sc\times\Ac\rightarrow \mathbb{R}^d$, if for any $h\in[H]$, there exist $d$ unknown measures $\mub_h^\ast:=[{\mu_h^\ast}^{(1)},\ldots,{\mu_h^\ast}^{(d)}]^\top$ over $\Sc$, and unknown vectors $\thetab_h^\ast,\gammab_h^\ast\in\mathbb{R}^d$ such that 
 $\Pb_h(.|s,a)=\left\langle \mub_h^\ast(.), \phib(s,a)\right\rangle$, $r_h(s,a)=\left\langle \thetab_h^\ast, \phib(s,a)\right\rangle$, and $c_h(s,a)=\left\langle \gammab_h^\ast,\phib(s,a)\right\rangle$.
\end{myassum}
This assumption highlights the definition of linear MDP, in which the Markov transition model, the reward functions, and the cost functions are linear in
a feature mapping $\phib$.
\subsection{Related works}\label{sec:relatedwork}
{\bf Safe RL with randomized policies:} The problem of Safe RL formulated with Constrained Markov Decision Process (CMDP) with a focus on unknown dynamics and \emph{randomized} policies is studied in \cite{efroni2020exploration,turchetta2020safe,garcelon2020conservative,zheng2020constrained,ding2020provably,qiu2020upper,ding2020natural,xu2020primal,kalagarla2020sample}. In the above-mentioned papers, the goal is to find the optimal randomized policy that maximizes the reward value function $V_r^\pi(s)$ (expected total reward) while ensuring the cost value function $V_c^\pi(s)$ (expected total cost) does not exceed a certain threshold. This safety requirement is defined over a \emph{horizon}, in expectation with respect to the environment and the randomization of the policy, and consequently is less strict than the safety requirement considered in this paper, which must be satisfied at each time-step an action is played.
In addition to their different problem formulations, the theoretical guarantees of these works fundamentally differ from the ones provided in our paper. The recent closely-related work of \cite{ding2020provably} studies constrained finite-horizon MDPs with a linear structure as considered in our paper via a primal-dual-type policy optimization
algorithm that achieves a $\Oc(dH^{2.5}\sqrt{T})$ regret and constraint violation and can only be applied to settings with finite action set $\Ac$.
The algorithm of \cite{efroni2020exploration} obtains a $\Oc(|\Sc|H^2\sqrt{|\Sc||\Ac|T})$ regret and constraint violation in the episodic finite-horizon tabular setting via linear program and primal-dual policy optimization. In \cite{qiu2020upper}, the authors study an adversarial stochastic shortest path problem under
constraints with $\Oc(|\Sc|H\sqrt{|\Ac|T})$ regret and constraint violation. \cite{ding2020natural} proposes a primal-dual
algorithm for solving discounted infinite horizon
CMDPs that achieves a global convergence with rate $\Oc(1/\sqrt{T})$ regarding both the optimality gap and the constraint violation. In contrast to the aforementioned works which can only guarantee bounds on the number of constraint violation, our algorithms \emph{never} violate the safety constraint during the learning process. 

Besides primal-dual methods, in \cite{chow2018lyapunov} Lyapunov functions are leveraged to handle the constraints. \cite{yu2019convergent} proposes a constrained
policy gradient algorithm with convergence guarantee. Both above-stated works focus on solving CMDPs with known transition model and constraint function without providing regret guarantees.


{\bf Safe RL with GPs and deterministic transition model and policies:} In another line of work, \cite{turchetta2016safe,berkenkamp2017safe,wachi2018safe,wachi2020safe} use Gaussian processes to model the dynamics with deterministic transitions and/or the value function in order to be able to estimate the constraints
and guarantee safe learning. Despite the fact that some of these algorithms are approximately safe, analysing the convergence is challenging and the regret analysis is lacking.
\section{Safe Linear UCB Q/V Iteration}\label{sec:SUCB}
In this section, we present \emph{Safe Linear Upper Confidence Bound Q/V Iteration} (\SUCB) summarized in Algorithm \ref{alg:SUCB}, which is followed by a high-level description of its performance in Section \ref{sec:SUCB}. First, we introduce the following necessary assumption and set of notations used in describing Algorithm \ref{alg:SUCB} and its analysis in the next sections.
\begin{myassum}[Non-empty safe sets]\label{assum:nonempty} For all $s\in\Sc$, there exists a known safe action $a_0(s)$ such that $a_0(s)\in \Ah^{{\rm safe}}(s)$ with known safety measure $\tau_h(s):=\left\langle \phib\left(s,a_0\left(s\right)\right),\gammab_h^\ast\right\rangle < \tau$ for all $h\in[H]$ .
\end{myassum}
Knowing safe actions $a_0(s)$ is necessary for solving the safe linear MDP setting studied in this paper, which requires the constraint \eqref{eq:safetyconstraintdeterministic} to be satisfied from the very first round. This assumption is also realistic in many practical examples, where the known safe action could be the one suggested by the current strategy of the company or a very cost-neutral action that does not necessarily have high reward but its cost is far from the threshold. It is possible to relax the assumption of knowing the cost of the safe actions $\tau_h(s)$. In this case, the agent starts by playing $a_0(s)$ for $T_h(s)$ rounds at time-steps $h$ in order to construct a conservative estimator for the gap $\tau-\tau_h(s)$. $T_h(s)$ is selected in an adaptive way and in Appendix \ref{sec:unknowntauhs}, we show that $\frac{16\log(K)}{(\tau-\tau_h(s))^2}  \leq T_h(s)\leq \frac{64\log(K)}{(\tau-\tau_h(s))^2}$. 
After $T_h(s)$ rounds, the agent relies on these estimates of $\tau_h(s)$ in the computation of estimated safe set of policies (discussed shortly).

\paragraph{Notations.} For any vector $\x\in\mathbb{R}^d$, define the normalized vector $\tilde\x : =\frac{\x}{\norm{\x}_2}$. We define the span of the safe feature $\phib\left(s,a_0\left(s\right)\right)$ as  $\Vc_s={\rm span}\left(\phib\left(s,a_0\left(s\right)\right)\right):=\left\{\alpha\phib\left(s,a_0\left(s\right)\right):\alpha\in\mathbb{R}\right\}$ and the orthogonal complement of $\Vc_s$ as $\Vc_s^\bot:=\{\y\in\mathbb{R}^d:\langle\y,\x\rangle=0,~\forall\x\in\Vc_s\}$. For any $\x\in \mathbb{R}^{d}$, denote by $\Phi_0(s,\x) := \left\langle \x,\tilde\phib\left(s,a_0\left(s\right)\right) \right\rangle \tilde\phib\left(s,a_0\left(s\right)\right) $ its projection on $\Vc_s$, and, by $\Phi_0^\bot(s,\x) := \x - \Phi_0(s,\x)$ its projection onto the orthogonal subspace $\Vc_s^\bot$. Moreover, for ease of notation, let $\phib_{h}^k:=\phib(s_h^k,a_h^k)$.

\begin{algorithm}[t]
\DontPrintSemicolon
\KwInput{$\Ac$, $\la$, $\delta$, $H$, $K$, $\tau$, $\kappa_h(s)$}
$\A_h^1 = \la I$, $\A_{h,s}^1 = \la \left(I-\tilde\phib\left(s,a_0\left(s\right)\right){\tilde\phib^\top\left(s,a_0\left(s\right)\right)}\right)$ $\bb_h^1=\rb_{h,s}^1=\mathbf{0},~\forall(s,h)\in\Sc\times[H], Q_{H+1}^k(.,.)=0,~\forall k\in[K]$
   \For{{\rm episodes} $k=1,\ldots,K$}
   {
   Observe the initial state $s_1^k$.\;
      \For{{\rm time-steps} $h=H,\ldots,1$}{ \label{line:firstbegin}
      Compute $\Ac_{h}^k(s)$ as in \eqref{eq:estimatedsafe} $\forall s\in \Sc$ . \label{line:safesetcomputation}\;
       Compute $Q^k_{h}(s,a)$ as in \eqref{eq:Q_h^k} $\forall (s,a)\in\Sc\times \Ahk(.)$.
      } \label{line:firstend}
      \For{ {\rm time-steps} $h=1,\ldots,H$}{ \label{line:secondbegin}
      Play $a_h^k=\argmax_{a\in \Ahk(s_h^k)}Q_h^{k}(s_h^k,a)$ and observe $s_{h+1}^k$, $r_h^k$ and $z_h^k$. \label{line:actionselection}
      }\label{line:secondend}
   }
 \caption{\SUCB}
   \label{alg:SUCB}
\end{algorithm}

\subsection{Overview}  
From a high-level point of
view, our algorithm is the safe version of LSVI-UCB proposed by \cite{jin2020provably}. In particular, each episode
consists of two loops over all time-steps. The first loop (Lines \ref{line:firstbegin}-\ref{line:firstend}) updates the quantities $\Ac_h^k$, estimated safe sets, and $Q_h^k$, action-value function, that are used to execute the \emph{upper confidence bound} policy $a_h^k=\argmax_{a\in \Ahk(s_h^k)}Q_h^{k}(s_h^k,a)$ in the second loop (Lines \ref{line:secondbegin}-\ref{line:secondend}). The key difference between \SUCB~and LSVI-UCB is the requirement that chosen actions $a_h^k$ must always belong to unknown safe sets $\Ac_h^{\rm safe}(s_h^k)$. To this end, at each episode $k\in[K]$, in an extra step in the first loop (Line \ref{line:safesetcomputation}), the agent computes a set $\Ac_h^{k}(s)$ for all $s\in\Sc$, which we will show is guaranteed to be a subset of the unknown safe set $\Ac_h^{\rm safe}(s)$, and therefore, is a good candidate to select action $a_h^k$ from in the second loop (Line \ref{line:actionselection}). Construction of $\Ac_h^{k}(s)$ depends on an appropriate confidence set around the unknown parameter $\gammab_h^\ast$ used in the definition of safety constraints (see Assumption \ref{assum:linearMDP}). Since the agent has knowledge of $\tau_h(s):=\left\langle \phib\left(s,a_0\left(s\right)\right),\gammab_h^\ast\right\rangle$ (see Assumption \ref{assum:nonempty}), 
it can compute $z_{h,s}^k:=\left\langle\Phi_0^\bot\left(s,\phib_h^k\right),\Phi_0^\bot\left(s,\gammab_h^\ast\right)\right\rangle+\epsilon_h^k= z_{h}^k - \frac{\left\langle {\phib}_{h}^k,\tilde\phib\left(s,a_0\left(s\right)\right)\right\rangle}{\norm{\phib\left(s,a_0\left(s\right)\right)}_2} \tau_h(s)$, i.e., the cost incurred by $a_h^k$  along the subspace $\Vc_s^\bot$, which is orthogonal to $\phib\left(s,a_0\left(s\right)\right)$. Thus, the agent does not need to build confidence sets around $\gammab_h^\ast$ along the normalized safe feature vector, $\tilde\phib\left(s,a_0\left(s\right)\right)$. Instead, it only builds the following confidence sets around $\Phi_0^\bot\left(s,\gammab_h^\ast\right)$ which is along the orthogonal direction of $\tilde\phib\left(s,a_0\left(s\right)\right)$:
\begin{align}\label{eq:gammaconfidenceset}
    \Cc_{h}^k(s):=\left\{\boldsymbol \nu \in \mathbb{R}^{d}: \norm{\boldsymbol \nu-\gammab_{h,s}^k}_{\A_{h,s}^k} \leq \beta\right\},
\end{align}
where $\gammab_{h,s}^k:=\left(\A_{h,s}^k\right)^{-1}\rb_{h,s}^k$ is the regularized least-squares estimator of $\Phi_0^\bot\left(s,\gammab_h^\ast\right)$ computed by the inverse of Gram matrix $\A_{h,s}^k:= \la\left(I-\tilde\phib\left(s,a_0\left(s\right)\right){\tilde\phib^\top\left(s,a_0\left(s\right)\right)}\right)+\sum_{j=1}^{k-1}\Phi_0^\bot\left(s,\phib_h^j\right)\Phi_0^{\bot,\top}\left(s,\phib_h^j\right)$ and $\rb_{h,s}^k:=\sum_{j=1}^{k-1}z_{h,s}^j\Phi_0^\bot\left(s,\phib_h^j\right)$.
The exploration factor $\beta$ will be defined shortly in Theorem~\ref{thm:SUCBregret} such that it guarantees that the event 
\begin{align}\label{eq:event1}
    \Ec_1:=\left\{\Phi_0^\bot\left(s,\gammab_h^\ast\right)\in\Cc_{h}^k(s),~\forall(s,h,k)\in\Sc\times[H]\times[K]\right\}
\end{align}
i.e., $\Phi_0^\bot\left(s,\gammab_h^\ast\right)$ belongs to the confidence sets $\Cc_h^k(s)$, holds with high probability. In the implementations, we treat $\beta$ as a tuning parameter.
Conditioned on event $\Ec_1$, the agent is ready to compute the following inner approximations of the true unknown safe sets $\Ac_h^{\rm safe}$ for all $s\in\Sc$:

\begin{align}\label{eq:estimatedsafe}
 &\Ahk(s)=\left\{a\in\Ac:\frac{\left\langle \Phi_0\left(s,\phib(s,a)\right),\tilde\phib\left(s,a_0\left(s\right)\right)\right\rangle}{\norm{\phib\left(s,a_0\left(s\right)\right)}_2} \tau_h(s)+\left\langle \gammab_{h,s}^k , \Phi_0^\bot\left(s,\phib(s,a)\right)\right\rangle+\beta\norm{\Phi_0^\bot\left(s,\phib(s,a)\right)}_{\left(\A_{h,s}^k\right)^{-1}}\leq \tau\right\}.
\end{align}

Note that $\frac{\left\langle \Phi_0\left(s,\phib(s,a)\right),\tilde\phib\left(s,a_0\left(s\right)\right)\right\rangle}{\norm{\phib\left(s,a_0\left(s\right)\right)}_2} \tau_h(s)$ is the known cost of action $a$ at state $s$ along direction $\tilde\phib\left(s,a_0\left(s\right)\right)$ and $\max_{\nub\in\Cc_h^k(s)}\left\langle\Phi_0^\bot\left(s,\phib(s,a)\right),\nub\right\rangle =\left\langle \gammab_{h,s}^k , \Phi_0^\bot\left(s,\phib(s,a)\right)\right\rangle+\beta\norm{\Phi_0^\bot\left(s,\phib(s,a)\right)}_{\left(\A_{h,s}^k\right)^{-1}}$ is its maximum possible cost in the orthogonal space $\Vc_s^\bot$. Thus, $\frac{\left\langle \Phi_0\left(s,\phib(s,a)\right),\tilde\phib\left(s,a_0\left(s\right)\right)\right\rangle}{\norm{\phib\left(s,a_0\left(s\right)\right)}_2} \tau_h(s)+\left\langle \gammab_{h,s}^k , \Phi_0^\bot\left(s,\phib(s,a)\right)\right\rangle+\beta\norm{\Phi_0^\bot\left(s,\phib(s,a)\right)}_{\left(\A_{h,s}^k\right)^{-1}}$ is a high probability upper bound on the true unknown cost $\langle \phib(s,a),\gammab_h^\ast\rangle$, which implies that $\Ahk(s)\subset\Ac_h^{\rm safe}(s)$. 

\begin{proposition}\label{prop:safe} Conditioned on $\Ec_1$ in \eqref{eq:event1}, for all $(s,h,k)\in\Sc\times[H]\times[K]$, it holds that $\left\langle \phib(s,a),\gammab_{h}^{\ast}\right \rangle \leq \tau,~\forall a\in\Ahk(s)$. 
\end{proposition}

Thus, conditioned on $\Ec_1$, the decision rule $a_h^k:=\argmax_{a\in \Ahk(s_h^k)}Q_h^{k}(s_h^k,a)$ in Line \ref{line:actionselection} of Algorithm \ref{alg:SUCB} suggests that $a_h^k$ does not violate the safety constraint. Note that $\Ahk(s)$ is always non-empty, since as a consequence of Assumption \ref{assum:nonempty}, the safe action $a_0(s)$ is always in $\Ahk(s)$.

Now that the estimated safe sets $\Ahk(s)$ are constructed, we describe how the action-value functions $Q_h^k$ are computed to be used in the UCB decision rule, selecting the action $a_h^k$ in the second loop of the algorithm. The linear structure of the MDP allows us to parametrize $Q_h^\ast(s,a)$ by a linear form $\langle\w_h^\ast,\phib(s,a)\rangle$, where $\w_h^\ast :=\thetab_h^\ast+\int_{\Sc} V_{h+1}^\ast(s^\prime)d\mub(s^\prime)$. Thus, a natural idea to estimate $Q_h^\ast(s,a)$ is to solve least-squares problem for $\w_h^\ast$. In fact, for all $(s,a)\in\Sc\times\Ahk(.)$, the agent computes $Q^k_h(s,a)$ defined as
\begin{align} \label{eq:Q_h^k}
    Q^k_h(s,a)=&\min\left\{\left\langle \w_h^k , \phib(s,a)\right\rangle+\kappa_h(s)\beta\norm{\phib(s,a)}_{\left(\A_h^{k}\right)^{-1}},
    H\right\},
\end{align}
where $\w_h^k:=\left(\A_h^{k}\right)^{-1}\bb_{h}^k$ is the regularized least-squares estimator of $\w^\ast_h$ computed by the inverse of Gram matrix $\A_h^k:=\la I+\sum_{j=1}^{k-1}\phib_{h}^j{\phib_{h}^j}^\top$ and
    $\bb_h^k:=\sum_{j=1}^{k-1} \phib_{h}^j \left[r_h^j+\max_{a\in\Ac_{h+1}^k(s_{h+1}^j)}Q_{h+1}^k(s_{h+1}^j,a)\right]$.
Here, $\kappa_h(s)\beta\norm{\phib(s,a)}_{\left(\A_h^{k}\right)^{-1}}$ is an exploration bonus that is characterized by: 1) $\beta$ that encourages enough exploration regarding the uncertainty about $r$ and $\mathbb{P}$; and 2) $\kappa_h(s)>1$ that encourages enough exploration regarding the uncertainty about $c$. While we make use of standard analysis of unsafe bandits and MDPs \cite{abbasi2011improved} and \cite{jin2020provably} to define $\beta$, appropriately quantifying $\kappa_h(s)$ is the main challenge the presence of safety constraints brings to the analysis of \SUCB~compared to the unsafe LSVI-UCB and it is stated in Lemma \ref{lemm:optimism}.
\section{Theoretical guarantees of \SUCB}\label{sec:theoreticalguarantees}
In this section, we discuss the technical challenges the presence of safety constraints brings to our analysis and provide a regret bound for \SUCB. Before these, we make the remaining necessary assumptions under which our proposed algorithm operates and achieves good
regret bound. 

\begin{myassum}[Subgaussian Noise]\label{assum:noise} For all $(h,k)\in[H]\times[K]$, $\epsilon_{h}^k$ is a zero-mean $\sigma$-subGaussian random variable.
\end{myassum}

\begin{myassum}[Boundedness]\label{assum:boundedness} Without loss of generality, $\norm{\phib(s,a)}_2\leq 1$ for all $(s,a)\in \Sc\times\Ac$, and $\max\left(\norm{\mub_h^\ast(\Sc)}_2,\norm{\thetab^\ast_h}_2,\norm{\gammab^\ast_h}_2\right)\leq \sqrt{d}$ for all $h\in[H]$. 
\end{myassum}

\begin{myassum}[Star convex sets]\label{assum:startconvexity} For all $s\in\Sc$, the set $\Dc(s):=\left\{\phib(s,a):a\in\Ac\right\}$ is a star convex set around the safe feature $\phib\left(s,a_0\left(s\right)\right)$, i.e., for all $\x\in\Dc(s)$ and $\alpha\in[0,1]$, $\alpha \x+(1-\alpha)\phib\left(s,a_0\left(s\right)\right)\in\Dc(s)$.
\end{myassum}

Assumptions \ref{assum:noise} and \ref{assum:boundedness} are standard in linear MDP and bandit literature \cite{jin2020provably,pacchiano2020stochastic,amani2019linear}. Assumption \ref{assum:startconvexity}
is necessary to ensure that the agent has the opportunity to explore the feature space around the given safe feature vector $\phib\left(s,a_0\left(s\right)\right)$. For example, consider a simple setting where $\Sc=\{s_1\}, \Ac=\{a_1,a_2\}, H=1, \mub^\ast(s_1)=(1,1), \thetab^\ast = (0,1), \gammab^\ast = (0,1), \tau=2$, $a_0(s_1)=a_2$, and $\Dc(s_1)=\{\phib(s_1,a_1),\phib(s_1,a_2)\}=\{(0,1),(1,0)\}$, which is not a star convex set. Here, both actions $a_1$ and $a_2$ are safe. The optimal safe policy always plays $a_1$, which gives the highest reward. However, if $\Dc(s_1)$ does not contain the whole line connecting $(1,0)$ and $(0,1)$, the agent keeps playing $a_2$ and will not be able to explore other safe action and identify that the optimal policy would always select $a_1$. Also, it is worth mentioning that the star convexity of the sets $\Dc(s)$ is a milder assumption than convexity assumption considered in existing safe algorithms of \cite{amani2019linear,moradipari2019safe}.

Given these assumptions, we are now ready to present the formal guarantees of \SUCB~in the following theorem.

\begin{theorem}[Regret of \SUCB]\label{thm:SUCBregret}  Under Assumptions \ref{assum:linearMDP}, \ref{assum:nonempty}, \ref{assum:noise}, \ref{assum:boundedness}, and \ref{assum:startconvexity}, there exists an absolute constant $c_\beta>0$ such that for any fixed $\delta\in(0,0.5)$, if we set $\beta:= \max\left(\sigma\sqrt{d\log\left(\frac{2+\frac{2T}{\la}}{\delta}\right)}+\sqrt{\la d},c_\beta dH\sqrt{\log(\frac{dT}{\delta})}\right)$, and $\kappa_h(s):=\frac{2H}{\tau-\tau_h(s)}+1$, then with probability at least $1-2\delta$, it holds that
     $R_K\leq2H\sqrt{T\log(\frac{dT}{\delta})}+(1+\kappa)\beta\sqrt{2dHT\log\left(1+\frac{K}{d\la}\right)}$, where $\kappa:=\max_{(s,h)\in\Sc\times[H]}\kappa_h(s)$
\end{theorem}
Here, $T = KH$ is the total number of action plays. We observe that the regret  bound  is  of  the  same  order as that of state-of-the-art unsafe algorithms, such as LSVI-UCB \cite{jin2020provably}, with only an additional factor $\kappa$ in its second term. The complete proof is reported in the Appendix \ref{sec:proofofmaintheorem}. In the following section, we give a sketch of the proof.
\subsection{Proof sketch of Theorem \ref{thm:SUCBregret}}\label{sec:proofsketch}

First, we state the following theorem borrowed from \cite{abbasi2011improved,jin2020provably}.

\begin{theorem}[Thm.~2 in \cite{abbasi2011improved} and Lemma~B.4 in \cite{jin2020provably}]\label{thm:events}
For any fixed policy $\pi$, define 
$ V_h^k(s):=\max_{a\in\Ahk(s,a)}Q_h^k(s,a)$,
and the event
\begin{align}
    &\hspace{-15pt}\Ec_2:=\left\{\left| \langle\w_h^k,\phib(s,a)\rangle-Q_h^\pi(s,a)+[\mathbb{P}_h(V^\pi_{h+1}-V^k_{h+1})](s,a)\right|\leq \beta\norm{\phib(s,a)}_{\left(\A_h^{k}\right)^{-1}},\forall(a,s,h,k)\in\Ac\times\Sc\times[H]\times[K]\right\}\nn,
\end{align} 
and recall the definition of $\Ec_1$ in \eqref{eq:event1}. Then, under Assumptions \ref{assum:linearMDP}, \ref{assum:nonempty}, \ref{assum:noise}, \ref{assum:boundedness}, and the definition of $\beta$ in Theorem \ref{thm:SUCBregret}, there exists an absolute constant $c_\beta>0$, such that for any fixed $\delta\in(0,0.5)$, with probability at least $1-\delta$, the event $\Ec:=\Ec_2\cap\Ec_1$ holds.
\end{theorem}

As our main technical contribution, in Lemma \ref{lemm:optimism}, we prove that when $\kappa_h(s):=\frac{2H}{\tau-\tau_h(s)}+1$, then \emph{optimism in the face of safety constraint}, i.e., $ Q_h^\ast(s,a) \leq Q_h^k(s,a)$ is guaranteed. Intuitively, this is required because the maximization in Line \ref{line:actionselection} of Algorithm \ref{alg:SUCB} is not over the entire $\Ac^{\rm safe}_h(s_h^k)$, but only a subset of it. Thus, larger values of $\kappa_h(s)$ (compared to $\kappa_h(s)=1$ in unsafe algorithm LSVI-UCB) are needed to provide enough exploration to the algorithm so that the selected actions in $\Ahk(s_h^k)$ are -often enough- \emph{optimistic}, i.e., $Q_h^\ast(s,a) \leq Q_h^k(s,a)$.

\begin{lemma}[Optimism in the face of safety constraint in \SUCB]\label{lemm:optimism} Let $\kappa_h(s):=\frac{2H}{\tau-\tau_h(s)}+1$ and Assumptions \ref{assum:linearMDP},\ref{assum:nonempty},\ref{assum:noise},\ref{assum:boundedness},\ref{assum:startconvexity} hold. Then, conditioned on $\Ec$, it holds that $V_h^\ast(s)\leq V_h^k(s),\forall(s,h,k)\in\Sc\times[H]\times[K]$.
\end{lemma}
We report the proof in Appendix \ref{sec:proofofoptimism}. As a direct conclusion of Lemma \ref{lemm:optimism} and on event $\Ec_2$ defined in Theorem  \ref{thm:events}, we have 
\begin{align}
  Q_h^\ast(s,a) &\leq \left\langle\w_h^k,\phib(s,a)\right\rangle+\beta\norm{\phib(s,a)}_{\left(\A_h^{k}\right)^{-1}}\nn+[\mathbb{P}_h V^\ast_{h+1}- V^k_{h+1}](s,a)\tag{Event $\Ec_2$}\\
   &\leq Q_h^k(s,a).\tag{Lemma \ref{lemm:optimism}}
\end{align}
This is encapsulated in the following corollary.
\begin{corollary}[UCB]\label{corr:UCB}
Let $\kappa_h(s):=\frac{2H}{\tau-\tau_h(s)}+1$ and Let Assumptions \ref{assum:linearMDP},\ref{assum:nonempty},\ref{assum:noise},\ref{assum:boundedness},\ref{assum:startconvexity} hold. Then, conditioned on $\Ec$, it holds that $Q_h^\ast(s,a)\leq Q_h^k(s,a),\forall(a,s,h,k)\in\Ac\times\Sc\times[H]\times[K]$.
\end{corollary}
After proving UCB nature of \SUCB~using Lemma \ref{lemm:optimism}, we are ready to exploit the standard analysis of classical unsafe LSVI-UCB \cite{jin2020provably} to complete the analysis and establish the final regret bound of \SUCB.

\section{Extension to randomized policy selection}\label{sec:randomizedSUCB}
\SUCB~presented in Section \ref{sec:SUCB} can only output a deterministic policy. In this section, we show that
our results can be extended to the setting of randomized policy selection, which might be desirable in practice. 
A randomized policy $\pi:\Sc\times [H]\rightarrow \Delta_\Ac$ maps states and time-steps to distributions over actions such that $a\sim\pi(s,h)$ is the action the policy $\pi$ suggests the agent to play at time-step $h\in[H]$ when being at state $s\in\Sc$. At each episode $k$ and time-step $h\in[H]$, when being in state $s_h^k$, the agent must draw its action $a_h^k$ from a \emph{safe} policy $\pi_k(s_h^k,h)$ such that
\begin{align}\label{eq:safetyconstraintrandomized}
  \mathbb {E}_{a_h^k\sim\pi_k(s_h^k,h)}c_h(s_h^k,a_h^k)\leq \tau  
\end{align}
with high probability.
We accordingly define the \emph{unknown} set of safe policies by
\begin{align}
    \tilde \Pi^{\rm safe}:=\left\{\pi:\pi(s,h)\in\Gamma^{\rm safe}_h(s),~\forall(s,h)\in \Sc\times[H]\right\},\nn
\end{align}
where $\Gamma^{\rm safe}_h(s):=\left\{\theta\in\Delta_{\Ac}:\mathbb{E}_{a\sim\theta}c_h(s,a)\leq \tau\right\}$.
Thus, after observing state $s_h^k$ at time-step $h\in[H]$ in episode $k$, the agent's choice of policy must belong to $\Gamma^{\rm safe}_h(s_h^k)$ with high probability. In this formulation, the expectation in the definition of (action-) value functions for a policy $\pi$ is over both the environment and the randomness of policy $\pi$. We denote them by $\tilde V_h^\pi$ and $\tilde Q_h^\pi$ to distinguish them from $V_h^\pi$ and $Q_h^\pi$ defined in \eqref{eq:valuefunction} and \eqref{eq:actionstatevaluefunction} for a deterministic policy $\pi$. Let $\pi_\ast$ be the optimal safe policy such that $\tilde V^{\pi_\ast}_h(s):=\tilde V^{\ast}_h(s) = \sup_{\pi\in\tilde \Pi^{\rm safe}}\tilde V_h^\pi(s)$ for all $(s,h)\in\Sc\times [H]$. Thus, for all $(a,s,h)\in\Ac\times\Sc\times [H]$, the Bellman equations  for a safe policy $\pi\in\tilde \Pi^{\rm safe}$ and the optimal safe policy are
\begin{align}
    \tilde Q_h^\pi(s,a)=r_h(s,a)+[\mathbb{P}_h\tilde V^\pi_{h+1}](s,a),\quad
    \tilde V_h^\pi(s)=\mathbb{E}_{a\sim\pi(s,h)}\left[\tilde Q_h^\pi(s,a)\right],\label{eq:bellmanforpirandomized}\\
    \tilde Q_h^\ast(s,a)=r_h(s,a)+[\mathbb{P}_h\tilde V^\ast_{h+1}](s,a),\quad
     \tilde V_h^\ast(s)=\max_{\theta\in\Gamma^{\rm safe}_h(s)}\mathbb{E}_{a\in\theta}\left[\tilde Q_h^\ast(s,a)\right]\label{eq:bellmanforoptimalrandomized},
\end{align}
where $\tilde V_{H+1}^\pi(s)=\tilde V_{H+1}^\ast(s)=0$, and the cumulative regret is defined as $R_K:=\sum_{k=1}^K \tilde V_1^{\ast}(s_1^k)-\tilde V_1^{\pi_k}(s_1^k)$. This definition of safety constraint in \eqref{eq:safetyconstraintrandomized} frees us from star-convexity assumption on the sets $\Dc(s):=\left\{\phib(s,a):a\in\Ac\right\}$ (Assumption \ref{assum:startconvexity}), which is necessary for the deterministic policy selection approach. We propose a modification of \SUCB~which is tailored to this new formulation and termed Randomized \SUCB~(\RSUCB). This new algorithm also achieves a sub-linear regret with the same order as that of \SUCB, i.e., $\Otilde\left(\kappa\sqrt{d^3H^3T}\right)$.


While \RSUCB~respects a milder definition of the safety constraint (cf. \eqref{eq:safetyconstraintrandomized}) compared to that considered in \SUCB~(cf. \eqref{eq:safetyconstraintdeterministic}), it still possesses significant superiorities  over other existing algorithms solving CMDP with randomized policy selection \cite{efroni2020exploration,turchetta2020safe,garcelon2020conservative,zheng2020constrained,ding2020provably,qiu2020upper,ding2020natural,xu2020primal,kalagarla2020sample}. First, the safety constraint considered in these algorithms is defined by the \emph{cumulative} expected cost over a horizon falling below a certain threshold, while \RSUCB~guarantees that the expected cost incurred at each time-step an action is played (not over a horizon) is less than a threshold. Second, even for this looser definition of safety constraint, the best these algorithms can guarantee in terms of constraint satisfaction is a sub-linear bound on the number of constraint violation, whereas \RSUCB~ensures \emph{no constraint violation}.
\subsection{Randomized \SUCB}
We now describe \RSUCB~summarized in Algorithm~\ref{alg:SUCBrandomized}. Let $\phib^\theta(s):=\mathbb{E}_{a\sim\theta}\phib(s,a)$. At each episode $k\in[K]$, in the first loop, the agent computes the estimated set of true unknown set $\Gamma^{\rm safe}_h(s)$ for all $s\in\Sc$ as follows:

\begin{align}
    \Gamma^k_h(s)&:=\left\{\theta\in \Delta_{\Ac}:\mathbb{E}_{a\sim\theta}\left[\frac{\left\langle \Phi_0\left(s,\phib(s,a)\right),\tilde\phib\left(s,a_0\left(s\right)\right)\right\rangle}{\norm{\phib\left(s,a_0\left(s\right)\right)}_2} \tau_h(s)\right]+\max_{\nub\in\Cc_h^k(s)}\left\langle\Phi_0^\bot\left(s,\mathbb{E}_{a\sim\theta}\left[\phib(s,a)\right]\right),\nub\right\rangle\leq \tau\right\}\nn\\
    &=\left\{\theta\in\Delta_{\Ac}:\frac{\left\langle \Phi_0\left(s,\phib^\theta(s)\right),\tilde\phib\left(s,a_0\left(s\right)\right)\right\rangle}{\norm{\phib\left(s,a_0\left(s\right)\right)}_2} \tau_h(s)+\left\langle \gammab_{h,s}^k , \Phi_0^\bot\left(s,\phib^\theta(s)\right)\right\rangle+\beta\norm{\Phi_0^\bot\left(s,\phib^\theta(s)\right)}_{\left(\A_{h,s}^k\right)^{-1}}\leq\tau\right\}\label{eq:estimatedsafesetrandomized}.
\end{align}
Note that due to the linear structure of the MDP, we can again parametrize $\tilde Q_h^\ast(s,a)$ by a linear form $\langle\tilde\w_h^\ast,\phib(s,a)\rangle$, where $\tilde\w_h^\ast :=\thetab_h^\ast+\int_{\Sc}\tilde V_{h+1}^\ast(s^\prime)d\mub(s^\prime)$. In the next step, for all $(s,a)\in\Sc\times\Ac$, the agent computes
\begin{align} \label{eq:Q_h^krandomized}
    \tilde Q^k_h(s,a)&=\left\langle \tilde \w_h^k , \phib(s,a)\right\rangle+\kappa_h(s)\beta\norm{\phib(s,a)}_{\left(\A_h^{k}\right)^{-1}},
\end{align}
where $\tilde\w_h^k:=\left(\A_h^{k}\right)^{-1}\tilde\bb_{h}^k$ is the regularized least-squares estimator of $\tilde\w^\ast_h$ computed by the Gram matrix $\A_h^k$ and
    $\tilde \bb_h^k:=\sum_{j=1}^{k-1} \phib_{h}^j \left[r_h^j+\min\left\{\max_{\theta\in\Gamma^k_{h+1}(s_{h+1}^j)}\mathbb{E}_{a\sim\theta}\left[\tilde Q_{h+1}^k(s_{h+1}^j,a)\right],H\right\}\right]$.
After these computations in the first loop, the agent draws actions $a_h^k$ from distribution $\Gamma_h^k(s_h^k)$ in the second loop.
Define $\tilde V_h^k(s):=\min\left\{\max_{\theta\in \Gamma^k_h(s)}\mathbb{E}_{a\sim\theta}\left[\tilde Q_{h}^k(s,a)\right],H\right\}$, and 
\begin{align}
    &\Ec_3:=\left\{\left| \langle\tilde\w_h^k,\phib(s,a)\rangle-\tilde Q_h^\pi(s,a)+[\mathbb{P}_h\tilde V^\pi_{h+1}-\tilde V^k_{h+1}](s,a)\right|\leq \beta\norm{\phib(s,a)}_{\left(\A_h^{k}\right)^{-1}},\forall(a,s,h,k)\in\Ac\times\Sc\times[H]\times[K]\right\}\nn.
\end{align} 
It can be easily shown that the results stated in Theorem \ref{thm:events} hold for the settings focusing on randomized policies, i.e., under Assumptions \ref{assum:linearMDP}, \ref{assum:nonempty}, \ref{assum:noise}, and \ref{assum:boundedness}, and by the definition of $\beta$ in Theorem \ref{thm:SUCBregret}, with probability at least $1-2\delta$, the event $\tilde \Ec:=\Ec_1\cap\Ec_3$ holds.
Therefore, as a direct conclusion of Proposition \ref{prop:safe}, it is guaranteed that conditioned on $\Ec_1$, all the policies inside $\Gamma^k_h(s)$ are safe, i.e., $\Gamma^k_h(s)\subset \Gamma^{\rm safe}_h(s)$. Now, in the following lemma, we quantify $\kappa_h(s)$.

\begin{algorithm}[t]
\DontPrintSemicolon
\KwInput{$\Ac$, $\la$, $\delta$, $H$, $K$, $\tau$, $\kappa_h(s)$}
$\A_h^1 = \la I$, $\A_{h,s}^1 = \la \left(I-\tilde\phib\left(s,a_0\left(s\right)\right){\tilde\phib^\top\left(s,a_0\left(s\right)\right)}\right)$ $\tilde\bb_h^1=\rb_{h,s}^1=\mathbf{0},~\forall(s,h)\in\Sc\times[H], \tilde Q_{H+1}^k(.,.)=0,~\forall k\in[K]$\;
   \For{{\rm episodes} $k=1,\ldots,K$}{
  Observe the initial state $s_1^k$.\;
      \For{{\rm time-steps} $h=H,\ldots,1$}{
      Compute $\Gamma^k_h(s)$ as in \eqref{eq:estimatedsafesetrandomized} $\forall s\in\Sc$.\;
       Compute $\tilde Q_h^k(s,a)$ as in \eqref{eq:Q_h^krandomized} $\forall(s,a)\in\Sc\times\Ac$.
       }
      \For{ {\rm time-steps} $h=1,\ldots,H$}{
      Play $a_h^k\sim\argmax_{\theta\in \Gamma^k_h(s_h^k)}\mathbb{E}_{a\sim\theta}\left[\tilde Q_h^k(s_h^k,a)\right]$ and observe $s_{h+1}^k$, $r_h^k$ and $z_h^k$.
      }
  }
   \caption{\RSUCB}
   \label{alg:SUCBrandomized}
\end{algorithm}

\begin{lemma}[Optimism in the face of safety constraint in \RSUCB] \label{lemm:optimismrandomized} Let $\kappa_h(s):=\frac{2H}{\tau-\tau_h(s)}+1$ and Assumptions \ref{assum:linearMDP},\ref{assum:nonempty},\ref{assum:noise},\ref{assum:boundedness} hold. Then, conditioned on event $\tilde\Ec$, it holds that $\tilde V_h^\ast(s)\leq \tilde V_h^k(s),\forall(s,h,k)\in\Sc\times[H]\times[K]$.
\end{lemma}
The proof is included in Appendix \ref{sec:proofofoptimismrandomized}. Using Lemma \ref{lemm:optimismrandomized}, we show that $\tilde Q^\ast_h(s,a)\leq \tilde Q^k_h(s,a),\forall(a,s,h,k)\in\Ac\times\Sc\times[H]\times[K]$. This highlights the UCB nature of \RSUCB, allowing us to exploit the standard analysis of unsafe LSVI-UCB \cite{jin2020provably} to establish the regret bound.
\begin{theorem}[Regret of \RSUCB]\label{thm:randomizedSUCBregret}
Under Assumptions \ref{assum:linearMDP}, \ref{assum:nonempty}, \ref{assum:noise}, and \ref{assum:boundedness}, there exists an absolute constant $c_\beta>0$ such that for any fixed $\delta\in(0,1/3)$, and the definition of $\beta$ in Theorem \ref{thm:SUCBregret}, if we set $\kappa_h(s):=\frac{2H}{\tau-\tau_h(s)}+1$, then with probability at least $1-3\delta$, it holds that
     $R_K\leq2H\sqrt{T\log(\frac{dT}{\delta})}+2(1+\kappa)\beta\sqrt{2dHT\log\left(1+\frac{K}{d\la}\right)}$, where $\kappa:=\max_{(s,h)\in\Sc\times[H]}\kappa_h(s)$.
\end{theorem}
See Appendix \ref{sec:proofofmaintheoremrandomized} for the proof.
\begin{figure}[t!]
\centering
\begin{tikzpicture}
\node at (0,0) {\includegraphics[scale=0.3]{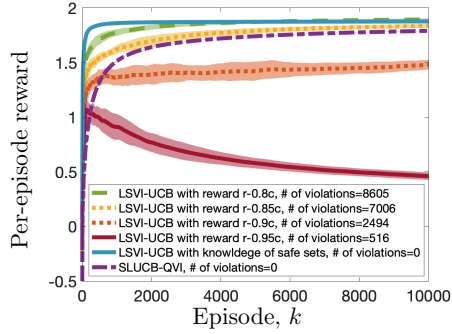}};
\node at (-3,0) [rotate=90,scale=0.9]{Per-episode reward};
\node at (0,-2.23) [scale=0.9]{Episode, $k$};
\end{tikzpicture}
\caption{Comparison of \SUCB~to the unsafe state-of-the-art verifying that: 1) when LSVI-UCB \cite{jin2020provably} has knowledge of $\gammab^\ast_h$, it outperforms \SUCB~(without knowledge of $\gammab^\ast_h$) as expected; 2) when LSVI-UCB does not know $\gammab^\ast_h$ (as is the case for \SUCB) and its goal is to maximize $r-\lambda^\prime c$ instead of $r$, larger $\lambda^\prime$ leads to smaller per-episode reward and number of constraint violations while the number of constraint violations for \SUCB~is zero.}
\label{fig:baseline}
\end{figure}

\begin{figure*}[t!]
\centering
\begin{subfigure}{2.2in}
\centering
\begin{tikzpicture}
\node at (0,0) {\includegraphics[scale=0.24]{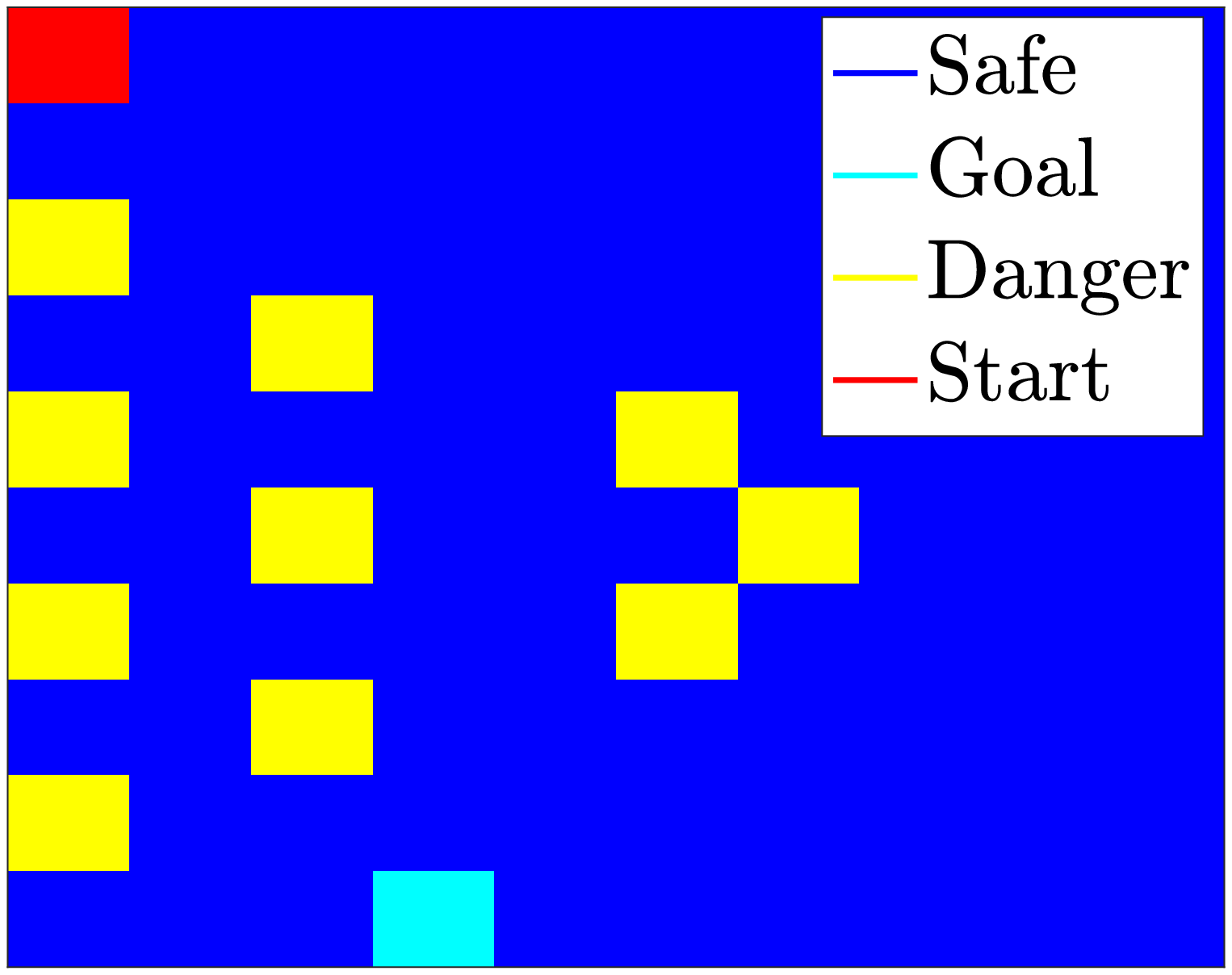}};
\node at (-2.3,0) [rotate=90,scale=0.9]{ };
\node at (0,-1.76) [scale=0.9]{ };
\end{tikzpicture}
\caption{Map}
\label{fig:map}
\end{subfigure}
\begin{subfigure}{2.2in}
\centering
\begin{tikzpicture}
\node at (0,0) {\includegraphics[scale=0.24]{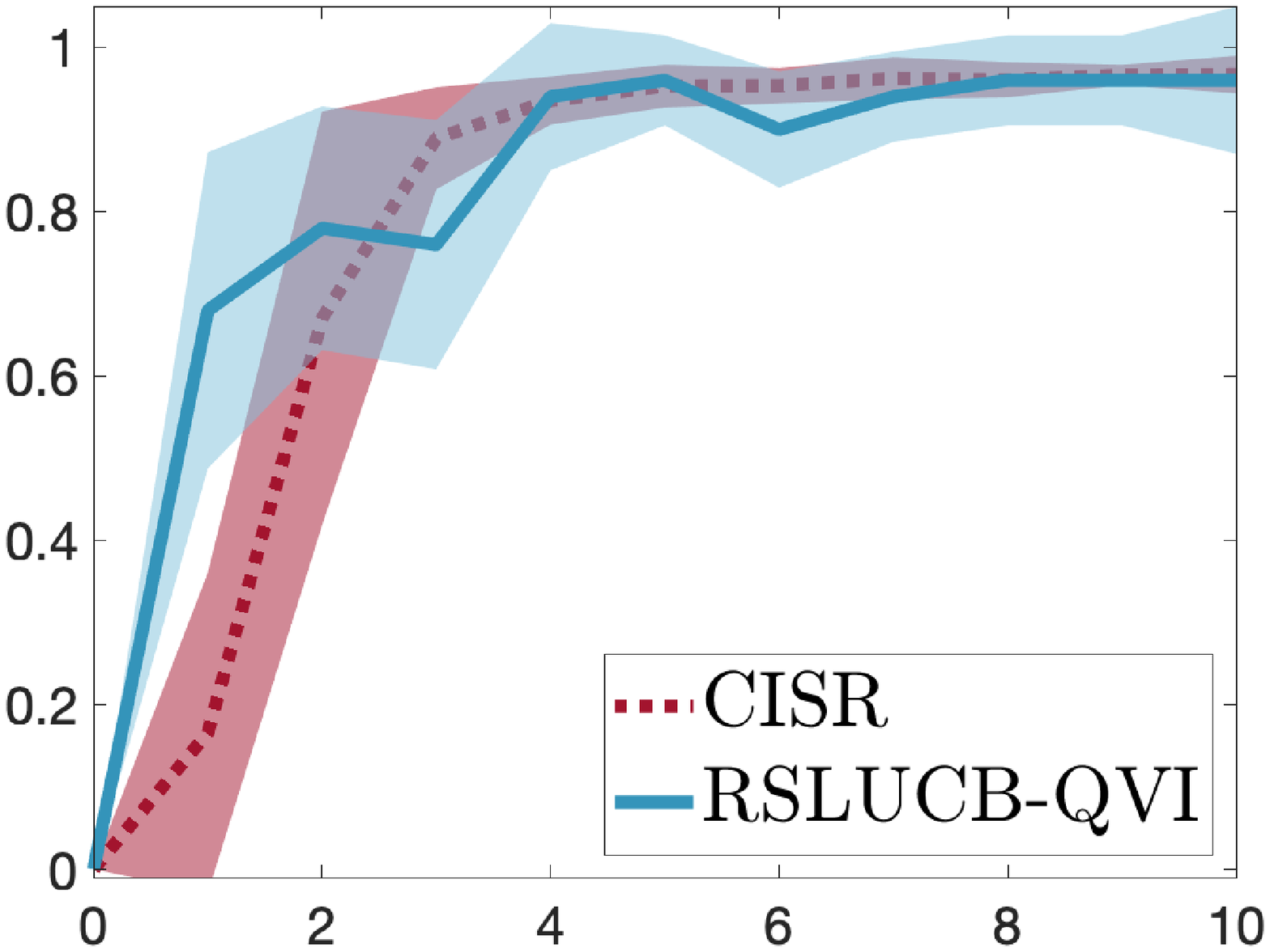}};
\node at (-2.3,0) [rotate=90,scale=0.9]{Success Rate};
\node at (0,-1.76) [scale=0.9]{Interaction unit};
\end{tikzpicture}
\caption{Success rate}
\label{fig:successrate}
\end{subfigure}
\centering
\begin{subfigure}{2.2in}
\centering
\begin{tikzpicture}
\node at (0,0) {\includegraphics[scale=0.24]{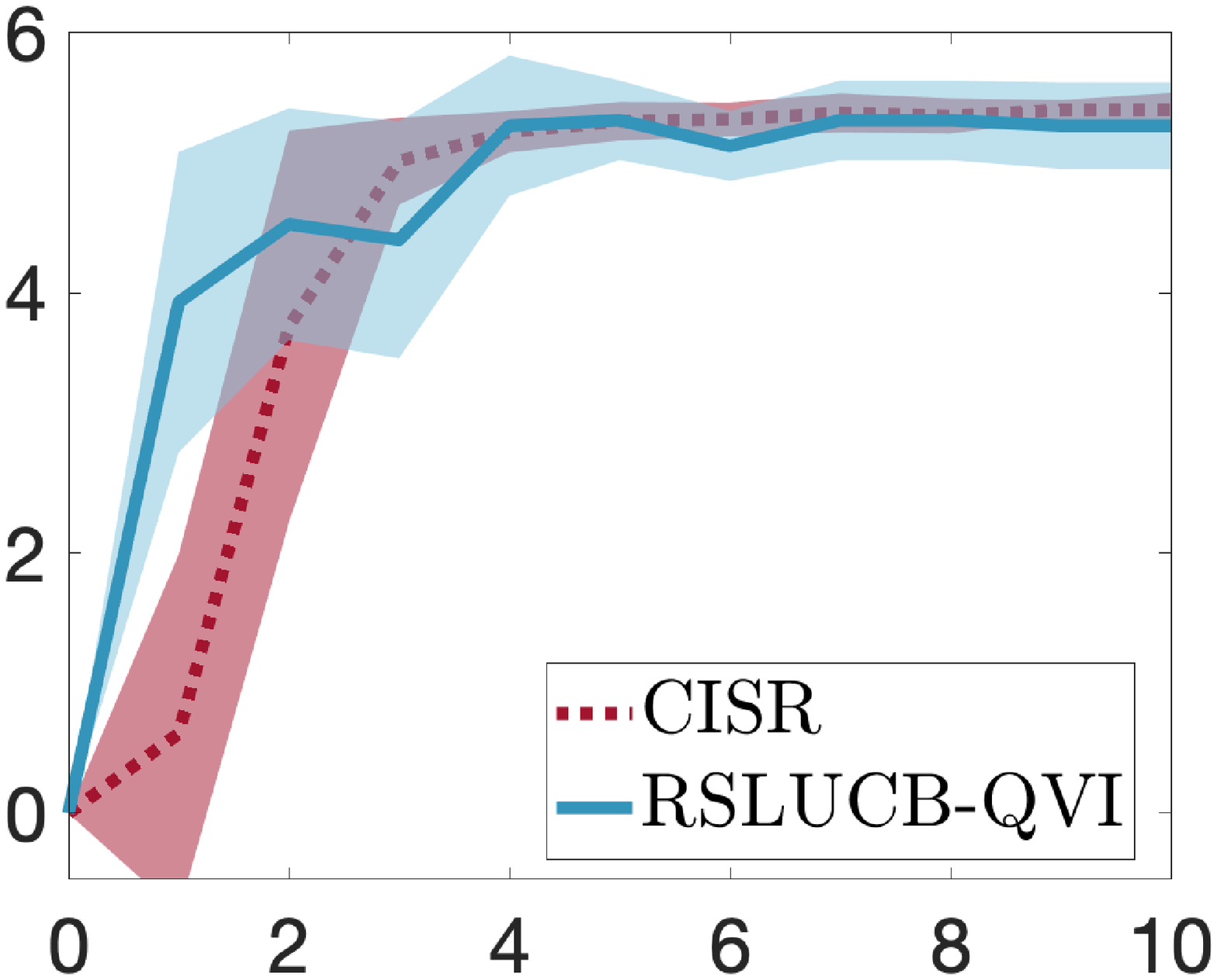}};
\node at (-2.3,0) [rotate=90,scale=0.9]{Average Return};
\node at (0,-1.76) [scale=0.9]{Interaction unit};
\end{tikzpicture}
\caption{Average return}
\label{fig:averagereturn}
\end{subfigure}
\caption{Comparison of \RSUCB~ and CISR \cite{turchetta2020safe} in Frozen Lake environment.}
\label{fig:frozenlake}
\end{figure*}

\section{Experiments}
In this section, we present numerical simulations \footnote{All the experiments are implemented in Matlab on a 2020 MacBook Pro with 32GB of RAM.} to complement and confirm our theoretical findings. We evaluate the performance of \SUCB~on synthetic environments and implement \RSUCB~on the \emph{Frozen Lake} environment from OpenAI Gym \cite{brockman2016openai}.

\subsection{\SUCB~on synthetic environments} 
The results shown in Figure \ref{fig:baseline} depict averages over 20 realizations, for which we have chosen $\delta=0.01$, $\sigma=0.01$, $\la=1$, $d = 5$, $\tau=0.5$, $H=3$ and $K=10000$. The parameters $\{\thetab_h^\ast\}_{h\in[H]}$ and $\{\gammab_h^\ast\}_{h\in[H]}$ are drawn from $\mathcal{N}(0,I_{d})$. In order to tune parameters $\{\mub_h^\ast(.)\}_{h\in[H]}$ and the feature map $\phib$ such that they are compatible with Assumption \ref{assum:linearMDP}, we consider that the feature space $\{\phib(s,a):(s,a)\in\Sc\times\Ac\}$ is a subset
of the $d$-dimensional simplex and $\e_i^\top\mub_h^\ast(.)$ is an arbitrary probability measure over $\Sc$ for all $i\in[d]$. This guarantees that Assumption \ref{assum:linearMDP} holds. 

Computing safe sets $\Ahk(s)$ in the first loop of \SUCB~(Line \ref{line:safesetcomputation}), is followed by selecting an action that maximizes a linear function (in feature map $\phib$) over the feature space $\Dc_h^k(s_h^k):=\left\{\phib(s_h^k,a):a\in\Ahk(s_h^k)\right\}$ in its second loop (Line \ref{line:actionselection}). Unfortunately,
even if the feature space $\{\phib(s,a):(s,a)\in\Sc\times\Ac\}$ is convex, the set $\Dc_h^k(s_h^k)$ can have a form over which maximizing the linear function is intractable. In our experiments, we define map $\phib$ such that the sets $\Dc(s)$ are star convex and \emph{finite} around $\phib\left(s,a_0\left(s\right)\right)$ with $N=100$ (see Definition \ref{def:finitestarconvex}) and therefore, we can show that the optimization problem in Line \ref{line:actionselection} of \SUCB~can be solved efficiently (see Appendix \ref{sec:tractablety} for a proof).

\begin{definition}[Finite star convex set]\label{def:finitestarconvex}
A star convex set $\Dc$ around $\x_0\in\mathbb{R}^d$ is finite,
if there exist finitely many vectors $\{\x_i\}_{i=1}^N$ such that $\Dc=\cup_{i=1}^N[\x_0,\x_i]$, where $[\x_0,\x_i]$ is the line connecting $\x_0$ and $\x_i$.
\end{definition}
Figure \ref{fig:baseline} depicts the average per-episode reward of \SUCB~and compares it to that of baseline and emphasizes the value of \SUCB~in terms of respecting the safety constraints at all time-steps. Specifically, we compare \SUCB~with 1) LSVI-UCB \cite{jin2020provably} when it has knowledge of safety constraints, i.e., $\gammab^\ast_h$; and 2) LSVI-UCB, when it does not know $\gammab^\ast_h$ (as is the case for \SUCB) and its goal is to maximize the function $r-\lambda^\prime c$, with the constraint being pushed into the objective function, for different values of $\lambda^\prime=0.8, 0.85, 0.9$ and $0.95$. Thus, playing costly actions is discouraged via low rewards. The plot verifies that LSVI-UCB with knowledge of $\gammab^\ast_h$ outperforms \SUCB~without knowledge of $\gammab^\ast_h$ as expected. Also, larger $\lambda^\prime$ leads to smaller per-episode reward and number of constraint violations when LSVI-UCB seeks to maximize $r-\lambda^\prime c$ (without knowledge of $\gammab^\ast_h$)  while the number of constraint violations for \SUCB~is zero.

\subsection{\RSUCB~on Frozen Lake environment}
We evaluate the performance of \RSUCB~in the Frozen Lake environment. The agent seeks to reach a goal in a $10\times 10$ 2D map (Figure \ref{fig:map}) while avoiding dangers. At each time step, the agent can move in four directions, i.e., $\Ac=\{a_1:\text{left}, a_2:\text{right}, a_3:\text{down}, a_4:\text{up}\}$. 
With probability $0.9$ it moves in the desired direction and with probability $0.05$ it moves in either of the orthogonal directions. We set $H=1000$, $K=10$, $d=|\Sc|=100$, and $\mub^\ast(s)\sim \Nc(0,I_d)$ for all $s\in\Sc=\{s_1,\ldots,s_{100}\}$. We then properly specified the feature map $\phib(s,a)$ for all $(s,a)\in \Sc\times\Ac$ by solving a set of linear equations such that the transition specifics of the environment explained above are respected. In order to interpret the requirement of avoiding dangers as a constraint of form \eqref{eq:safetyconstraintrandomized}, we tuned $\gammab^\ast$ and $\tau$ as follows: the cost of playing action $a\in\Ac$ at state $s\in\Sc$ is the probability of the agent moving to one of the danger states. Therefore a safe policy insures that the expected value of probability of moving to a danger state is a small value. To this end, we set $\gammab^\ast=\sum_{s\in {\rm Danger~states}}\mub^\ast(s)$ and $\tau=0.1$. Also, for each state $s\in\Sc$ a safe action, playing which leads to one of the danger states with small probability ($\tau=0.1$) is given to the agent.  We solve a set of linear equations to tune $\thetab^\ast$ such that at each state $s\in\Sc$, the direction which leads to a state that is closest to the goal state gives the agent a reward 1, while playing other three directions gives it a reward  0.01. This model persuades the agent to move towards to the goal.

After specifying the feature map $\phib$ and tuning all parameters, we implemented \RSUCB~for 10 interaction units (episodes) i.e, $K=10$) each consisting of 1000 time-steps (horizon), i.e., $H=1000$). During each interaction unit (episode) and after each move, the agent can end up in one of three kinds of states: 1) goal, resulting in a successful
termination of the interaction unit; 2) danger, resulting in a failure and the consequent termination of
the interaction unit; 3) safe. The agent receives a return of 6 for reaching the goal and 0.01 otherwise. 

In Figure \ref{fig:frozenlake}, we report the average of success rate and return over 20 agents for each of which we implemented \RSUCB~10 times and compare our results with that of CISR proposed by \cite{turchetta2020safe} in which a teacher helps the agent in selecting safe actions by making interventions. While the performances of both approaches, \RSUCB~and CISR, are fairly comparable, an important point to consider is that each interaction unit (episode) in CISR consists of 10000 time-steps whereas this number is 1000 in \RSUCB. Notably, the learning rate of \RSUCB~is faster than that of CISR. Also it is noteworthy that we compared\RSUCB~with CISR when it uses the \emph{optimized} intervention, which gives the best results compared to other types of intervention.

\section{Conclusion}

In this paper, we developed \SUCB~and \RSUCB, two safe RL algorithms in the setting of finite-horizon linear MDP. For these algorithms, we provided sub-linear regret bounds $\Otilde\left(\kappa\sqrt{d^3H^3T}\right)$, where $H$ is the duration of each episode, $d$ is the dimension of the feature mapping, $\kappa$ is a constant characterizing the safety constraints, and $T=KH$ is the total number of action plays. We proved that with high probability, they never violate the unknown safety constraints. Finally, we implemented \SUCB~and \RSUCB~on synthetic and Frozen Lake environments, respectively, which confirms that our algorithms have performances comparable to that of state-of-the-art that either have knowledge of the safety constraint or take advantage of a teacher's advice helping the agent avoid unsafe actions.
\section*{Acknowledgements}
We thank Mohammad Ghavamzadeh for discussion and bringing finite star convex sets into our attention.

\bibliographystyle{apalike}
\bibliography{main}
\newpage
\appendix
\onecolumn

\section{\SUCB~proofs}

In this section, we prove the technical statements in Sections \ref{sec:SUCB} and \ref{sec:theoreticalguarantees}. First, recall the definitions of the following events that we repeatedly refer to throughout this section:

\begin{align}\label{eq:event11}
    \Ec_1:=\left\{\Phi_0^\bot\left(s,\gammab_h^\ast\right)\in\Cc_{h}^k(s),~\forall(s,h,k)\in\Sc\times[H]\times[K]\right\},
\end{align}

\begin{align}\label{eq:event22}
    &\Ec_2:=\left\{\abs{ \langle\w_h^k,\phib(s,a)\rangle-Q_h^\pi(s,a)+[\mathbb{P}_hV^\pi_{h+1}-V^k_{h+1}](s,a)}\leq \beta\norm{\phib(s,a)}_{\left(\A_h^{k}\right)^{-1}},\forall(a,s,h,k)\in\Ac\times\Sc\times[H]\times[K]\right\}.
\end{align} 

\subsection{Proof of Proposition  \ref{prop:safe}}\label{sec:proofofproposition}
Let $a\in \Ahk(s)$. Recall that $\Phi_0(s,\x)=\left\langle \x,\tilde\phib\left(s,a_0\left(s\right)\right) \right\rangle \tilde\phib\left(s,a_0\left(s\right)\right)$ for any $\x\in\mathbb{R}^d$. By the definition of $\Ahk(s)$ in \eqref{eq:estimatedsafe}, we have
\begin{align}\label{eq:equ1}
    \frac{\left\langle \Phi_0\left(s,\phib(s,a)\right),\tilde\phib\left(s,a_0\left(s\right)\right)\right\rangle}{\norm{\phib\left(s,a_0\left(s\right)\right)}_2} \tau_h(s)+\left\langle \gammab_{h,s}^k , \Phi_0^\bot\left(s,\phib(s,a)\right)\right\rangle+\beta\norm{\Phi_0^\bot\left(s,\phib(s,a)\right)}_{\left(\A_{h,s}^k\right)^{-1}}\leq \tau
\end{align}
Moreover, using Cauchy-Schwarz inequality and conditioned on event $\Ec_1$ in \eqref{eq:event11}, we get
\begin{align}
    \left|\left\langle\gammab_{h,s}^k-\Phi_0^\bot\left(s,\gammab_h^\ast\right),\Phi_0^\bot\left(s,\phib(s,a)\right)\right\rangle\right|\leq \beta \norm{\Phi_0^\bot\left(s,\phib(s,a)\right)}_{\left(\A_{h,s}^k\right)^{-1}},\end{align}
    and thus,
    \begin{align}\label{eq:equ2}
    \left\langle\Phi_0^\bot\left(s,\gammab_h^\ast\right),\Phi_0^\bot\left(s,\phib(s,a)\right)\right\rangle\leq \left\langle\gammab_{h,s}^k,\Phi_0^\bot\left(s,\phib(s,a)\right)\right\rangle+\beta \norm{\Phi_0^\bot\left(s,\phib(s,a)\right)}_{\left(\A_{h,s}^k\right)^{-1}}.
\end{align}
Note that $ \langle \Phi_0^\bot\left(s,\gammab_h^\ast\right),\Phi_0^\bot\left(s,\phib(s,a)\right)\rangle = \left\langle \gammab_h^\ast,\phib(s,a)\right\rangle- \langle \gammab_{h,s}^\ast,{\Phi_0\left(s,\phib(s,a)\right)}\rangle = \langle\gammab_h^\ast,\phib(s,a)\rangle - \frac{\left\langle \Phi_0\left(s,\phib(s,a)\right),\tilde\phib\left(s,a_0\left(s\right)\right)\right\rangle}{\norm{\phib\left(s,a_0\left(s\right)\right)}_2} \tau_h(s)$. Combining this fact with \eqref{eq:equ1} and \eqref{eq:equ2} concludes that
\begin{align}
    \left\langle \gammab_h^\ast,\phib(s,a)\right\rangle&= \frac{\left\langle {\Phi_0\left(s,\phib(s,a)\right)},\tilde\phib\left(s,a_0\left(s\right)\right)\right\rangle}{\norm{\phib\left(s,a_0\left(s\right)\right)}_2} \tau_h(s)+\left\langle \Phi_0^\bot\left(s,\gammab_h^\ast\right),\Phi_0^\bot\left(s,\phib(s,a)\right)\right\rangle\nn\\
    &\leq\frac{\left\langle \Phi_0\left(s,\phib(s,a)\right),\tilde\phib\left(s,a_0\left(s\right)\right)\right\rangle}{\norm{\phib\left(s,a_0\left(s\right)\right)}_2} \tau_h(s)+\left\langle \gammab_{h,s}^k , \Phi_0^\bot\left(s,\phib(s,a)\right)\right\rangle+\beta\norm{\Phi_0^\bot\left(s,\phib(s,a)\right)}_{\left(\A_{h,s}^k\right)^{-1}}\tag{Eqn. \eqref{eq:equ2}}\\
    &\leq \tau \tag{Eqn. \eqref{eq:equ1}},
\end{align}
which implies that $a\in \Ahk(s)$, as desired.

\subsection{Proof of Lemma \ref{lemm:optimism}}\label{sec:proofofoptimism}
Before we start the main proof, we introduce vectors $\{\w_h^\pi\}_{h\in[H]}$ for any policy $\pi$:
\begin{align}\label{eq:whpi}
    \w_h^\pi :=\thetab_h^\ast+\int_{\Sc}V_{h+1}^\pi(s^\prime)d\mub(s^\prime).
\end{align}

From the Bellman equation in \eqref{eq:bellmanforpi} and the linearity of the MDP in Assumption \ref{assum:linearMDP}, we have:
\begin{align}
    Q_h^\pi(s,a):=\left\langle\phib(s,a),\w_h^\pi\right\rangle.
\end{align}
See Proposition 2.3 in \cite{jin2020provably} for the proof.

Now, we prove Lemma \ref{lemm:optimism} by induction. First, we prove the base case at time-step $H+1$. The statement holds because $V_{H+1}^\ast(s)=V_{H+1}^k(s)=0$. Now, suppose the statement holds for time-step $h+1$. We prove it also holds for time-step $h$. For all $(s,h,k)\in\Sc\times[H]\times[K]$, let 
\begin{align}
   a_h^k(s):=\argmax_{a\in\Ahk(s)}Q_h^k(s,a)\quad\text{and}\quad a_h^\ast(s):=\argmax_{a\in\Ah^{{\rm safe}}(s)}Q_h^\ast(s,a).
\end{align}
We consider the following two cases:

1) If $a_h^\ast(s)\in \Ahk(s)$, we have
\begin{align}
    V_h^k(s) = \max_{a\in\Ahk(s)}Q_h^k(s,a)&\geq Q_h^k(s,a_h^\ast(s))\nn\\
    &\geq Q_h^\ast(s,a_h^\ast(s))+\mathbb{E}_{s'\sim \Pb(.|s,a_h^\ast(s))}\left[V_{h+1}^k(s')-V_{h+1}^\ast(s')\right]\tag{Conditioned on $\Ec_2$ in \eqref{eq:event22}}\\
    &\geq Q_h^\ast(s,a_h^\ast(s))=V_h^\ast(s),\tag{Induction assumption}
\end{align}
as desired.

2) Now, we recall the definition of $\Ahk(s)$ in \eqref{eq:estimatedsafe} and focus on the other case when $a_h^\ast(s)\notin \Ahk(s)$, which means
\begin{align}\label{eq:mgreaterthanc}
 \frac{\left\langle \Phi_0\left(s,\phib(s,a^\ast_h(s))\right),\tilde\phib\left(s,a_0\left(s\right)\right)\right\rangle}{\norm{\phib\left(s,a_0\left(s\right)\right)}_2} \tau_h(s)
    +\left\langle \gammab_{h,s}^k , \Phi_0^\bot\left(s,\phib(s,a^\ast_h(s))\right)\right\rangle+\beta\norm{\Phi_0^\bot\left(s,\phib(s,a^\ast_h(s))\right)}_{\left(\A_{h,s}^k\right)^{-1}}> \tau.
\end{align}

Now, we observe that $a_0(s)\in\Ahk(s)$. Recall that $\tilde\phib\left(s,a_0\left(s\right)\right)=\frac{\phib\left(s,a_0\left(s\right)\right)}{\norm{\phib\left(s,a_0\left(s\right)\right)}_2}$ and note that $\Phi_0\left(s,\phib\left(s,a_0\left(s\right)\right)\right)=\phib\left(s,a_0\left(s\right)\right)$ and $\Phi_0^\bot\left(s,\phib(s,a_0(s))\right)=\mathbf{0}$. Thus
\begin{align}
    \frac{\left\langle \phib\left(s,a_0\left(s\right)\right),\tilde\phib\left(s,a_0\left(s\right)\right)\right\rangle}{\norm{\phib\left(s,a_0\left(s\right)\right)}_2} \tau_h(s)+\left\langle \gammab_{h,s}^k , \Phi_0^\bot\left(s,\phib(s,a_0(s))\right)\right\rangle+\beta\norm{\Phi_0^\bot\left(s,\phib(s,a_0(s))\right)}_{\left(\A_{h,s}^k\right)^{-1}}=\tau_h(s)<\tau,
\end{align}
which implies that $a_0(s)\in\Ahk(s)$ or equivalently $\phib\left(s,a_0\left(s\right)\right)\in 
   \Dc_h^k(s):=\left\{\phib(s,a):a\in\Ac_h^k(s)\right\}$. Now, let 
\begin{align}\label{eq:alpha}
    \alpha_h^k(s):=\max \left\{\alpha\in[0,1]: \alpha\phib(s,a^\ast_h(s))+(1-\alpha)\phib\left(s,a_0\left(s\right)\right)\in\Dc_h^k(s)\right\}.
\end{align}
Assumption \ref{assum:nonempty} guarantees that $\alpha_h^k(s)$ exists for all $(s,k)\in\Sc\times[H]\times[K]$. Note that $\Phi_0\left(s,\alpha\phib(s,a^\ast_h(s))+(1-\alpha)\phib\left(s,a_0\left(s\right)\right)\right) = \alpha\Phi_0\left(s,\phib(s,a^\ast_h(s))\right)+(1-\alpha)\phib\left(s,a_0\left(s\right)\right) $ and $\Phi_0^\bot\left(s,\alpha\phib(s,a^\ast_h(s))+(1-\alpha)\phib\left(s,a_0\left(s\right)\right)\right) = \alpha\Phi_0^\bot\left(s,\phib(s,a^\ast_h(s))\right)$. Thus, by the definition of $\Dc_h^k(s)$, we have
\begin{align}
    \alpha_h^k(s):=\max \Bigg\{\alpha\in[0,1]:& \frac{\left\langle \alpha\Phi_0\left(s,\phib(s,a^\ast_h(s))\right)+(1-\alpha)\phib\left(s,a_0\left(s\right)\right),\tilde\phib\left(s,a_0\left(s\right)\right)\right\rangle}{\norm{\phib\left(s,a_0\left(s\right)\right)}_2} \tau_h(s)\nn\\
    &+\alpha\left\langle \gammab_{h,s}^k , \Phi_0^\bot\left(s,\phib(s,a^\ast_h(s))\right)\right\rangle+\alpha\beta\norm{\Phi_0^\bot\left(s,\phib(s,a^\ast_h(s))\right)}_{\left(\A_{h,s}^k\right)^{-1}}\leq\tau\Bigg\}\label{eq:alphahk}.
\end{align}

For all $(s,k)\in\Sc\times[K]$, at time-step $h$, let $\y_{h}^k(s):=\alpha_h^k(s)\phib(s,a^\ast_h(s))+(1-\alpha_h^k(s))\phib\left(s,a_0\left(s\right)\right)$. Thus, the definition of $\alpha_h^k(s)$ in \eqref{eq:alpha} implies that $\y_h^k(s)\in\Dc_h^k(s)$, and thus
\begin{align}
   \max_{a\in\Ahk(s)}Q_h^k(s,a)&\geq \min\left\{\left\langle \w_h^k , \y_{h}^k(s)\right\rangle+\kappa_h(s)\beta\norm{\y_{h}^k(s)}_{\left(\A_h^{k}\right)^{-1}},H\right\}\nn\\
   &=\min\left\{\left\langle \w_h^k -\w_h^\ast, \y_{h}^k(s)\right\rangle+\left\langle \w_h^\ast , \y_{h}^k(s)\right\rangle+\kappa_h(s)\beta\norm{\y_{h}^k(s)}_{\left(\A_h^{k}\right)^{-1}},H\right\}\label{eq:firstt}.
\end{align}
Conditioned on event $\Ec_2$ in \eqref{eq:event22}, and by the induction assumption, we have
\begin{align}
   -\beta\norm{\y_{h}^k(s)}_{\left(\A_h^{k}\right)^{-1}}&\leq \left\langle \w_h^k-\w_h^\ast , \y_{h}^k(s)\right\rangle+\mathbb{E}_{s'\sim \Pb(.|s,a_h^\ast(s))}\left[V_{h+1}^\ast(s')-V_{h+1}^k(s')\right]\nn\\
   &\leq \left\langle \w_h^k-\w_h^\ast , \y_{h}^k(s)\right\rangle\label{eq:secondd}.
\end{align}
By combining \eqref{eq:firstt} and \eqref{eq:secondd}, we conclude that
\begin{align}
   \max_{a\in\Ahk(s)}Q_h^k(s,a)&\geq\min\left\{\left\langle \w_h^\ast , \y_{h}^k(s)\right\rangle+(\kappa_h(s)-1)\beta\norm{\y_{h}^k(s)}_{\left(\A_h^{k}\right)^{-1}},H\right\}\nn\\
   &\geq \min\left\{\alpha_h^k(s)\left\langle \w_h^\ast , \phib(s,a^\ast_h(s))\right\rangle + (1-\alpha_h^k(s))\left\langle \w_h^\ast , \phib\left(s,a_0\left(s\right)\right)\right\rangle+(\kappa_h(s)-1)\beta\norm{\Phi_0^\bot\left(s,\y_{h}^k(s)\right)}_{\left(\A_{h,s}^k\right)^{-1}},H\right\}\nn\\
   &\geq \min\left\{\alpha_h^k(s)\left(\left\langle \w_h^\ast , \phib(s,a^\ast_h(s))\right\rangle +(\kappa_h(s)-1)\beta\norm{\Phi_0^\bot\left(s,\phib(s,a^\ast_h(s))\right)}_{\left(\A_{h,s}^k\right)^{-1}}\right),H\right\}, \label{eq:maxQgeq}
\end{align}
where the second inequality holds because $\norm{\y_{h}^k(s)}_{\left(\A_h^{k}\right)^{-1}}\geq \norm{\Phi_0^\bot\left(s,\y_{h}^k(s)\right)}_{\left(\A_{h,s}^k\right)^{-1}}$ (see Lemma 3 in \cite{pacchiano2020stochastic} for a proof). The last inequality follows from the fact that $(1-\alpha_h^k(s))\left\langle\w_h^\ast,\phib\left(s,a_0\left(s\right)\right)\right\rangle\geq 0$ as the reward is always positive, i.e., $r_h(s,a)\in[0,1]$ for all $(s,a,h)\in\Sc\times\Ac\times[H]$.

Now, we show that $\alpha_{h}^k(s)\geq \frac{\tau-\tau_h(s)}{\tau-\tau_h(s)+2\beta\norm{\Phi_0^\bot\left(s,\phib(s,a^\ast_h(s))\right)}_{\left(\A_{h,s}^k\right)^{-1}}}$, which eventually leads to a proper value for $\kappa_h(s)>1$ that guarantees for all $(s,h,k)\in\Sc\times[H]\times[K]$ it holds that $V_h^\ast(s)\leq V_h^k(s)$ conditioned on $\Ec=\Ec_1\cap\Ec_2$. Definitions of $\alpha_h^k(s)$ in \eqref{eq:alphahk} and the estimated safe set $\Ahk(s)$ in \eqref{eq:estimatedsafe} imply that for all $(s,h,k)\in\Sc\times[H]\times[K]$, we have
\begin{align}\label{eq:somethinginthemiddle}
   &\frac{(1-\alpha_h^k(s))\left\langle\phib\left(s,a_0\left(s\right)\right),\tilde\phib\left(s,a_0\left(s\right)\right)\right\rangle}{\norm{\phib\left(s,a_0\left(s\right)\right)}_2} \tau_h(s)+\alpha_h^k(s)\left[\frac{\left\langle \Phi_0\left(s,\phib(s,a^\ast_h(s))\right),\tilde\phib\left(s,a_0\left(s\right)\right)\right\rangle}{\norm{\phib\left(s,a_0\left(s\right)\right)}_2} \tau_h(s)\right.\nn\\
   &+\left.\left\langle \gammab_{h,s}^k , \Phi_0^\bot\left(s,\phib(s,a^\ast_h(s))\right)\right\rangle+\beta\norm{\Phi_0^\bot\left(s,\phib(s,a^\ast_h(s))\right)}_{\left(\A_{h,s}^k\right)^{-1}}\right] = \tau.
\end{align}

Let $M = \frac{\left\langle \Phi_0\left(s,\phib(s,a^\ast_h(s))\right),\tilde\phib\left(s,a_0\left(s\right)\right)\right\rangle}{\norm{\phib\left(s,a_0\left(s\right)\right)}_2} \tau_h(s)+\left\langle \gammab_{h,s}^k , \Phi_0^\bot\left(s,\phib(s,a^\ast_h(s))\right)\right\rangle+\beta\norm{\Phi_0^\bot\left(s,\phib(s,a^\ast_h(s))\right)}_{\left(\A_{h,s}^k\right)^{-1}}$. Note that due to \eqref{eq:mgreaterthanc}, $M>\tau$, and recall that $\tilde\phib\left(s,a_0\left(s\right)\right) = \frac{\phib\left(s,a_0\left(s\right)\right)}{\norm{\phib\left(s,a_0\left(s\right)\right)}_2}$. Therefore, \eqref{eq:somethinginthemiddle} gives that
\begin{align}
    0<\alpha_{h}^k(s) = \frac{\tau-\tau_h(s)}{M-\tau_h(s)}<1.
\end{align}

In order to lower bound $\alpha_{h}^k(s)$ (upper bound $M$), we first rewrite $M$ as
\begin{align}
    M = &\frac{\left\langle \Phi_0\left(s,\phib(s,a^\ast_h(s))\right),\tilde\phib\left(s,a_0\left(s\right)\right)\right\rangle}{\norm{\phib\left(s,a_0\left(s\right)\right)}_2} \tau_h(s)+\left\langle  \gammab_{h}^{\ast}, \Phi_0^\bot\left(s,\phib(s,a^\ast_h(s))\right)\right\rangle\nn\\
    &+\left\langle \gammab_{h,s}^k -\gammab_{h}^{\ast}, \Phi_0^\bot\left(s,\phib(s,a^\ast_h(s))\right)\right\rangle+\beta\norm{\Phi_0^\bot\left(s,\phib(s,a^\ast_h(s))\right)}_{\left(\A_{h,s}^k\right)^{-1}}\label{eq:M},
\end{align}
and show that 
\begin{enumerate}[(a)] 
\item $\frac{\left\langle \Phi_0\left(s,\phib(s,a^\ast_h(s))\right),\tilde\phib\left(s,a_0\left(s\right)\right)\right\rangle}{\norm{\phib\left(s,a_0\left(s\right)\right)}_2} \tau_h(s)+\left\langle  \gammab_{h}^{\ast}, \Phi_0^\bot\left(s,\phib(s,a^\ast_h(s))\right)\right\rangle\leq \tau$ because

\begin{align}
\hspace{-20pt}\frac{\left\langle \Phi_0\left(s,\phib(s,a^\ast_h(s))\right),\tilde\phib\left(s,a_0\left(s\right)\right)\right\rangle}{\norm{\phib\left(s,a_0\left(s\right)\right)}_2} \tau_h(s)+\left\langle  \gammab_{h}^{\ast}, \Phi_0^\bot\left(s,\phib(s,a^\ast_h(s))\right)\right\rangle &=\left\langle\gammab_h^\ast,\left\langle  \Phi_0\left(s,\phib(s,a^\ast_h(s))\right),\tilde\phib\left(s,a_0\left(s\right)\right) \right\rangle \tilde\phib\left(s,a_0\left(s\right)\right) \right\rangle\nn\\
&+\left\langle\gammab_h^\ast,\Phi_0^\bot\left(s,\phib(s,a^\ast_h(s))\right)\right\rangle \nn\\
  &=\left\langle\gammab_h^\ast,\Phi_0\left(s,\phib(s,a^\ast_h(s))\right)\right\rangle+\left\langle\gammab_h^\ast,\Phi_0^\bot\left(s,\phib(s,a^\ast_h(s))\right)\right\rangle\nn\\
  &=\left\langle\gammab_h^\ast,\phib(s,a^\ast_h(s))\right\rangle\nn\\&\leq \tau. 
  \label{eq:leqtau}
\end{align}
\item $\left\langle \gammab_{h,s}^k -\gammab_{h}^{\ast}, \Phi_0^\bot\left(s,\phib(s,a^\ast_h(s))\right)\right\rangle\leq \beta\norm{\Phi_0^\bot\left(s,\phib(s,a^\ast_h(s))\right)}_{\left(\A_{h,s}^k\right)^{-1}}$, because conditioned on $\Ec_1$ in \eqref{eq:event11}, we have
\begin{align}
    \left\langle \gammab_{h,s}^k -\gammab_{h}^{\ast}, \Phi_0^\bot\left(s,\phib(s,a^\ast_h(s))\right)\right\rangle &= \left\langle \gammab_{h,s}^k -\Phi_0^\bot\left(s,\gammab_h^\ast\right), \Phi_0^\bot\left(s,\phib(s,a^\ast_h(s))\right)\right\rangle\nn\\
    &\leq \beta\norm{\Phi_0^\bot\left(s,\phib(s,a^\ast_h(s))\right)}_{\left(\A_{h,s}^k\right)^{-1}}.\label{eq:leqbeta}
\end{align}
\end{enumerate}
Now, we combine \eqref{eq:M}, \eqref{eq:leqtau} and \eqref{eq:leqbeta} to conclude that
\begin{align}
    M \leq \tau+2\beta\norm{\Phi_0^\bot\left(s,\phib(s,a^\ast_h(s))\right)}_{\left(\A_{h,s}^k\right)^{-1}} \quad\Rightarrow\quad  \alpha_h^k(s)\geq \frac{\tau-\tau_h(s)}{\tau-\tau_h(s)+2\beta\norm{\Phi_0^\bot\left(s,\phib(s,a^\ast_h(s))\right)}_{\left(\A_{h,s}^k\right)^{-1}}}.
\end{align}
This lower bound on $\alpha_h^k(s)$ combined with \eqref{eq:maxQgeq} gives
\begin{align}
 \max_{a\in\Ahk(s)}Q_h^k(s,a)&\geq\min\left\{ \frac{(\tau-\tau_h(s))\left(\left\langle \w_h^\ast , \phib(s,a^\ast_h(s))\right\rangle +(\kappa_h(s)-1)\beta\norm{\Phi_0^\bot\left(s,\phib(s,a^\ast_h(s))\right)}_{\left(\A_{h,s}^k\right)^{-1}}\right)}{\tau-\tau_h(s)+2\beta\norm{\Phi_0^\bot\left(s,\phib(s,a^\ast_h(s))\right)}_{\left(\A_{h,s}^k\right)^{-1}}},H\right\}
\end{align}
Let $M_1=\beta\norm{\Phi_0^\bot\left(s,\phib(s,a^\ast_h(s))\right)}_{\left(\A_{h,s}^k\right)^{-1}}$. We observe that  Therefore $\max_{a\in\Ac_h^{\rm safe}}Q_h^\ast(s,a) = \max_{a\in\Ac_h^{\rm safe}}\min\left\{Q_h^\ast(s,a),H\right\} =\min\left\{\max_{a\in\Ac_h^{\rm safe}}Q_h^\ast(s,a),H\right\} $. Therefore
\begin{align}
    \max_{a\in\Ahk(s)}Q_h^k(s,a)\geq \max_{a\in\Ac_h^{\rm safe}}Q_h^\ast(s,a) &\iff \left(\tau-\tau_h(s)\right)\left(\max_{a\in\Ac_h^{\rm safe}}Q_h^\ast(s,a)+(\kappa_h(s)-1)M_1\right)\geq \left(\tau-\tau_h(s)+2M_1\right)\max_{a\in\Ac_h^{\rm safe}}Q_h^\ast(s,a)\nn\\
    &\iff \left(\tau-\tau_h(s)\right)(\kappa_h(s)-1)\geq2\max_{a\in\Ac_h^{\rm safe}}Q_h^\ast(s,a)\nn\\
    &\iff \left(\tau-\tau_h(s)\right)(\kappa_h(s)-1)\geq2H\nn\\
    &\iff \kappa_h(s)\geq\frac{2H}{\tau-\tau_h(s)}+1,
\end{align}
as desired.

\subsection{Proof of Theorem \ref{thm:SUCBregret}}\label{sec:proofofmaintheorem}
The key property of optimism in the face of safety constraint in \SUCB, which is proved in Appendix \ref{sec:proofofoptimism} as our main technical allows us to follow the standard steps in establishing the regret bound of unsafe LSVI-UCB in \cite{jin2020provably} to complete the proof of Theorem \ref{thm:SUCBregret}.

Conditioned on event $\Ec_2$ in \eqref{eq:event22}, for any $(a,s,h,k)\in\Ac\times\Sc\times[H]\times[K]$, we have
\begin{align}
    Q_h^k(s,a)-Q_h^{\pi_k}(s,a) &= \min\left\{\left\langle \w_h^k , \phib(s,a)\right\rangle+\kappa_h(s)\beta\norm{\phib(s,a)}_{\left(\A_h^{k}\right)^{-1}},H\right\}-Q_h^{\pi_k}(s,a)\nn\\
    &\leq \left\langle \w_h^k , \phib(s,a)\right\rangle+\kappa_h(s)\beta\norm{\phib(s,a)}_{\left(\A_h^{k}\right)^{-1}}-Q_h^{\pi_k}(s,a)
    \nn\\&\leq\mathbb{E}_{s'\sim \Pb(.|s,a)}\left[V_{h+1}^k(s')-V_{h+1}^{\pi_k}(s')\right]+\left(1+\kappa_h(s)\right)\beta\norm{\phib(s,a)}_{\left(\A_h^{k}\right)^{-1}}\label{eq:Q-Q}.
\end{align}
Let $\delta_h^k:=V_{h}^k(s_h^k)-V_{h}^{\pi_k}(s_h^k)$ and $\zeta_{h+1}^k:=\mathbb{E}_{s'\sim \Pb(.|s_h^k,a_h^k)}\left[V_{h+1}^k(s')-V_{h+1}^{\pi_k}(s')\right]-\delta_{h+1}^k$.
We can write
\begin{align}
    \delta_h^k&=V_{h}^k(s_h^k)-V_{h}^{\pi_k}(s_h^k)\nn\\
    &= Q_h^k(s_h^k,a_h^k)-Q_h^{\pi_k}(s_h^k,a_h^k)\nn\\
    &\leq \mathbb{E}_{s'\sim \Pb(.|s_h^k,a_h^k)}\left[V_{h+1}^k(s')-V_{h+1}^{\pi_k}(s')\right]+\left(1+\kappa_h(s)\right)\beta\norm{\phib_h^k}_{\left(\A_h^{k}\right)^{-1}} \tag{Eqn.~\eqref{eq:Q-Q}}\\
    &=\delta_{h+1}^k+\zeta_{h+1}^k+\left(1+\kappa_h(s)\right)\beta\norm{\phib_h^k}_{\left(\A_h^{k}\right)^{-1}}.
\end{align}
Now, conditioning on event $\Ec=\Ec_1\cap\Ec_2$, we bound the cumulative regret as follows:
\begin{align}
    R_K&=\sum_{k=1}^K V_1^{\ast}(s_1^k)-V_1^{\pi_k}(s_1^k)\leq \sum_{k=1}^K\delta_1^k\tag{Lemma \ref{lemm:optimism}}\\
    &\leq \sum_{k=1}^K\sum_{h=1}^H \zeta_{h}^k+\sum_{k=1}^K\sum_{h=1}^H\left(1+\kappa_h(s)\right)\beta\norm{\phib_h^k}_{\left(\A_h^{k}\right)^{-1}}\nn\\
    &\leq \sum_{k=1}^K\sum_{h=1}^H \zeta_{h}^k+\left(1+\kappa\right)\beta\sum_{k=1}^K\sum_{h=1}^H\norm{\phib_h^k}_{\left(\A_h^{k}\right)^{-1}}.\label{eq:laststepregret}
\end{align}
We observe that $\{\zeta_h^k\}$ is a martingale
difference sequence satisfying $|\zeta_h^k|\leq 2H$. Thus, thanks to Azuma-Hoeffding inequality, we have
\begin{align}
    \mathbb{P}\left(\sum_{k=1}^K\sum_{h=1}^H\zeta_{h}^k\leq 2H\sqrt{T\log(dT/\delta)}\right)\geq 1-\delta.\label{eq:azoma}
\end{align}
In order to bound $\sum_{k=1}^K\sum_{h=1}^H\norm{\phib_h^k}_{\left(\A_h^{k}\right)^{-1}}$, note that for any $h\in[H]$, we have
\begin{align}
    \sum_{k=1}^K\norm{\phib_h^k}_{\left(\A_h^{k}\right)^{-1}}&\leq \sqrt{K\sum_{k=1}^K\norm{\phib_h^k}^2_{\left(\A_h^{k}\right)^{-1}}}\tag{Cauchy-Schwartz inequality}\\
    &\leq\sqrt{2K\log\left(\frac{\det\left(\A_h^K\right)}{\det\left(\A_h^1\right)}\right)}\label{eq:firstine}\\
    &\leq\sqrt{2dK\log\left(1+\frac{K}{d\la}\right)}.\label{eq:secondine}
\end{align}
In inequality \eqref{eq:firstine}, we used the standard argument in regret analysis of linear bandits \cite{abbasi2011improved} (Lemma~11) as follows: 
\begin{align}\label{eq:standardarg}
    \sum_{t=1}^n \min\left(\norm{\y_t}^2_{\Vb_{t}^{-1}},1\right)\leq 2\log\frac{\det \Vb_{n+1}}{\det\Vb_{1}}\quad \text{where}\quad \Vb_n = \Vb_{1}+\sum_{t=1}^{n-1}\y_t\y^\top_t.
\end{align}
In inequality \eqref{eq:secondine}, we used Assumption \ref{assum:boundedness} and the fact that $\det(\A)=\prod_{i=1}^d \lambda_i(\A) \leq ({\rm trace}(\A)/d)^d$. Combining \eqref{eq:laststepregret}, \eqref{eq:azoma}, and \eqref{eq:secondine}, we have with probability at least $1-2\delta$
\begin{align}
    R_K\leq 2H\sqrt{T\log(dT/\delta)}+(1+\kappa)\beta\sqrt{2dHT\log\left(1+\frac{K}{d\la}\right)}.
\end{align}

\subsection{Unknown \texorpdfstring{$\tau_h(s)$}{k}}\label{sec:unknowntauhs}
In this section, we relax Assumption \ref{assum:nonempty}, and instead assume that we only have the knowledge of safe actions $a_0(s)$, and remove the assumption on the knowledge about their costs $\tau_h(s)$. Similar results are provided by \cite{pacchiano2020stochastic}.

Let $k$ be the number of times the agent has played action $a_0(s)$ at time-step $h$, and $\hat\tau_h(s)$ be the empirical mean estimator of $\tau_h(s)$. Then, for any $\delta\in(0,1)$, we have
\begin{align}
    \mathbb{P}\left(\tau_h(s)\leq\hat\tau_h(s)+\sqrt{2\log(1/\delta)/k}\right)\geq 1-\delta.
\end{align}
If we let $\delta=1/K^2$, then we have
\begin{align}
    \mathbb{P}\left(\abs{\hat\tau_h(s)-\tau_h(s)}\leq 2\sqrt{\log(K)/k},~\forall k\in[K]\right)\geq 1-2/K.
\end{align}
We find $T_h(s)$, the number of time the agent must play action $a_0(s)$ at state $s$ and time-step $h$ in an adaptive manner as follow. Let $T_h(s)$ be the first time that $\hat\tau_h(s)+6\sqrt{\log(K)/T_h(s)}\leq \tau$. Thus, we have
\begin{align}
    \tau_h(s)+4\sqrt{\log(K)/T_h(s)}\leq \tau \Rightarrow \frac{16\log(K)}{(\tau-\tau_h(s))^2}\leq T_h(s).
\end{align}

Note that in this case $4\sqrt{\log(K)/T_h(s)}$ is a conservative estimation for $\tau-\tau_h(s)$.

Now we show that it will not take much longer than $\frac{16\log(K)}{(\tau-\tau_h(s))^2}$ that this first time happens. Conversely, for any $k\geq \frac{64\log(K)}{(\tau-\tau_h(s))^2}$, we observe that

\begin{align}
    \hat\tau_h(s)+6\sqrt{\log(K)/k}\leq \tau_h(s)+8\sqrt{\log(K)/k}\leq \tau.
\end{align}  
Therefore, we conclude that
\begin{align}
  \frac{16\log(K)}{(\tau-\tau_h(s))^2}  \leq T_h(s)\leq \frac{64\log(K)}{(\tau-\tau_h(s))^2},
\end{align}
and $4\sqrt{\log(K)/T_h(s)}$ is a conservative estimate for $\tau-\tau_h(s)$.

\section{Randomized \SUCB~proofs}\label{sec:discussiononrandomized}
In this section, we prove the technical statements in Section \ref{sec:randomizedSUCB}. First, recall the definition of the following event that we repeatedly refer to throughout this section:

\begin{align}\label{eq:event3}
    \Ec_3:=\left\{\abs{ \langle\tilde\w_h^k,\phib(s,a)\rangle-\tilde Q_h^\pi(s,a)+[\mathbb{P}_h\tilde V^\pi_{h+1}-\tilde V^k_{h+1}](s,a)}\leq \beta\norm{\phib(s,a)}_{\left(\A_h^{k}\right)^{-1}},\forall(a,s,h,k)\in\Ac\times\Sc\times[H]\times[K]\right\}.
\end{align}
In the following theorem, we state that $\Ec_3$, focusing on randomized policy selection, is a high probability event.

\begin{theorem}[Thm.~2 in \cite{abbasi2011improved} and Lemma~B.4 in \cite{jin2020provably}]\label{thm:tildeevents}
Define 
\begin{align}
\tilde V_h^k(s):=\min\left\{\max_{\theta\in \Gamma^k_h(s)}\mathbb{E}_{a\sim\theta}\left[\tilde Q_{h}^k(s,a)\right],H\right\}
\end{align}
and recall the definition of $\Ec_1$ in \eqref{eq:event11}. Then, for any fixed policy $\pi$, under Assumptions \ref{assum:linearMDP}, \ref{assum:nonempty}, \ref{assum:noise}, and \ref{assum:boundedness}, and the definition of $\beta$ in Theorem \ref{thm:SUCBregret}, there exists an absolute constant $c_\beta>0$, such that for any fixed $\delta\in(0,0.5)$, with probability at least $1-2\delta$, the event $\tilde\Ec:=\Ec_1\cap\Ec_3$ holds.
\end{theorem}

\subsection{Proof of Lemma \ref{lemm:optimismrandomized}}\label{sec:proofofoptimismrandomized}
First, similar to vectors  $\{\w_h^\pi\}_{h\in[H]}$ in \eqref{eq:whpi} for deterministic policy selection setting, we introduce vectors $\{\tilde\w_h^\pi\}_{h\in[H]}$ for any policy $\pi$:
\begin{align}
    \tilde\w_h^\pi :=\thetab_h^\ast+\int_{\Sc}\tilde V_{h+1}^\pi(s^\prime)d\mub(s^\prime).
\end{align}
From the Bellman equation in \eqref{eq:bellmanforpirandomized} and the linearity of the MDP in Assumption \ref{assum:linearMDP}, we have:
\begin{align}
    \tilde Q_h^\pi(s,a):=\left\langle\phib(s,a),\tilde\w_h^\pi\right\rangle.
\end{align}

Now, similar to the proof of Lemma \ref{lemm:optimism}, we start proving this Lemma \ref{lemm:optimismrandomized} by induction. First, we prove the base case at time-step $H+1$. The statement holds because $\tilde V_{H+1}^\ast(s)= \tilde V_{H+1}^k(s)=0$. Now, suppose the statement holds for time-step $h+1$. We prove it also holds for time-step $h$. For all $(s,h,k)\in\Sc\times[H]\times[K]$, let 
\begin{align}
   \pi_k(s,h):=\argmax_{\theta\in \Gamma^k_h(s)}\mathbb{E}_{a\sim\theta}\left[\tilde Q_{h}^k(s,a)\right]\quad\text{and}\quad \pi_\ast(s,h):=\argmax_{\theta\in \Gamma^{\rm safe}_h(s)}\mathbb{E}_{a\sim\theta}\left[\tilde Q_{h}^\ast(s,a)\right].
\end{align}
We consider the following two cases:

1) If $\pi_\ast(s,h)\in \Gamma^k_h(s)$, we have
\begin{align}
    \tilde V_h^k(s) = \min\left\{\max_{\theta\in \Gamma^k_h(s)}\mathbb{E}_{a\sim\theta}\left[\tilde Q_{h}^k(s,a)\right],H\right\}&\geq \min\left\{\mathbb{E}_{a\sim\pi_\ast(s,h)}\left[\tilde Q_{h}^k(s,a)\right],H\right\}\nn\\
    &\geq \min\left\{\mathbb{E}_{a\sim\pi_\ast(s,h)}\left[\tilde Q_h^\ast(s,a)+\mathbb{E}_{s'\sim \Pb(.|s,a)}\left[\tilde V_{h+1}^k(s')-\tilde V_{h+1}^\ast(s')\right]\right],H\right\}\tag{Conditioned on $\Ec_3$ in \eqref{eq:event3}}\\
    &\geq\min\left\{\mathbb{E}_{a\sim\pi_\ast(s,h)}\left[\tilde Q_h^\ast(s,a)\right],H\right\},\tag{Induction assumption}\\
     &=\mathbb{E}_{a\sim\pi_\ast(s,h)}\left[\tilde Q_h^\ast(s,a)\right] = V_h^\ast(s).
\end{align}
as desired.

2) Now, we recall the definition of $\Gamma^k_h(s)$ in \eqref{eq:estimatedsafesetrandomized} and focus on the other case when $\pi_\ast(s,h)\notin \Gamma^k_h(s)$, which means
\begin{align}\label{eq:mgreaterthancrandomized}
 \frac{\left\langle \Phi_0\left(s,\phib^{\pi_\ast(s,h)}(s)\right),\tilde\phib\left(s,a_0\left(s\right)\right)\right\rangle}{\norm{\phib\left(s,a_0\left(s\right)\right)}_2} \tau_h(s)+\left\langle \gammab_{h,s}^k , \Phi_0^\bot\left(s,\phib^{\pi_\ast(s,h)}(s)\right)\right\rangle+\beta\norm{\Phi_0^\bot\left(s,\phib^{\pi_\ast(s,h)}(s)\right)}_{\left(\A_{h,s}^k\right)^{-1}}>\tau.
\end{align}

Let $\pi_0(s,h)$ be the policy that always selects $a_0(s)$ for all $(s,h)\in\Sc\times[H]$. Now, we observe that $\pi_0(s,h)\in\Gamma^k_h(s)$. Recall that $\tilde\phib\left(s,a_0\left(s\right)\right)=\frac{\phib\left(s,a_0\left(s\right)\right)}{\norm{\phib\left(s,a_0\left(s\right)\right)}_2}$ and note that $\Phi_0\left(s,\phib\left(s,a_0\left(s\right)\right)\right)=\phib\left(s,a_0\left(s\right)\right)$ and $\Phi_0^\bot\left(s,\phib(s,a_0(s))\right)=\mathbf{0}$. Thus
\begin{align}
&\frac{\left\langle \Phi_0\left(s,\phib^{\pi_0(s,h)}(s)\right),\tilde\phib\left(s,a_0\left(s\right)\right)\right\rangle}{\norm{\phib\left(s,a_0\left(s\right)\right)}_2} \tau_h(s)+\left\langle \gammab_{h,s}^k , \Phi_0^\bot\left(s,\phib^{\pi_0(s,h)}(s)\right)\right\rangle+\beta\norm{\Phi_0^\bot\left(s,\phib^{\pi_0(s,h)}(s)\right)}_{\left(\A_{h,s}^k\right)^{-1}}\nn\\
&=\frac{\left\langle \phib\left(s,a_0\left(s\right)\right),\tilde\phib\left(s,a_0\left(s\right)\right)\right\rangle}{\norm{\phib\left(s,a_0\left(s\right)\right)}_2} \tau_h(s)+\left\langle \gammab_{h,s}^k , \Phi_0^\bot\left(s,\phib(s,a_0(s))\right)\right\rangle+\beta\norm{\Phi_0^\bot\left(s,\phib(s,a_0(s))\right)}_{\left(\A_{h,s}^k\right)^{-1}}=\tau_h(s)<\tau,
\end{align}
which implies that $\pi_0(s,h)\in\Gamma^k_h(s)$. Now, let $\tilde\pi_k(s,h):=\alpha_h^k(s)\pi_\ast(s,h)+(1-\alpha_h^k(s))\pi_0(s,h)$, where
\begin{align}\label{eq:alpharandomized}
   \alpha_h^k(s):=\left\{\max \alpha\in[0,1]: \alpha\pi_\ast(s,h)+(1-\alpha)\pi_0(s,h)\in\Gamma^k_h(s)\right\}.
\end{align}

Let $\phib^\theta(s):=\mathbb{E}_{a\sim\theta}\phib(s,a)$. We observe that
\begin{align}
    \phib^{\tilde\pi_k(s,h)}(s)&=\alpha_h^k(s)\phib^{\pi_\ast(s,h)}(s)+(1-\alpha_h^k(s))\phib^{\pi_0(s,h)}(s) \nn\\
    &= \alpha_h^k(s)\phib^{\pi_\ast(s,h)}(s)+(1-\alpha_h^k(s))\phib\left(s,a_0\left(s\right)\right).
\end{align}

Since $\tilde\pi_k(s,h)\in\Gamma_h^k(s)$ (see the definition of $\alpha_h^k(s)$ in \eqref{eq:alpharandomized}), for all $(s,k)\in\Sc\times[K]$, at time-step $h$, we have
\begin{align}
\hspace{-30pt}
\tilde V_h^k(s) = \min\left\{\max_{\theta\in \Gamma^k_h(s)}\mathbb{E}_{a\sim\theta}\left[\tilde Q_{h}^k(s,a)\right],H\right\}&\geq \min\left\{\mathbb{E}_{a\sim\tilde\pi(s,h)}\left[\tilde Q_{h}^k(s,a)\right],H\right\}\\
&=\min\left\{\mathbb{E}_{a\sim\tilde\pi_k(s,h)}\left[\langle \tilde \w_h^k , \phib(s,a)\rangle+\kappa_h(s)\beta\norm{\phib(s,a)}_{\left(\A_h^k\right)^{-1}}\right],H\right\}\\
&\geq\min\left\{\langle \tilde \w_h^k , \phib^{\tilde\pi_k(s,h)}(s)\rangle+\kappa_h(s)\beta\norm{\phib^{\tilde\pi_k(s,h)}(s)}_{\left(\A_h^k\right)^{-1}},H\right\}\tag{Jensen's Inequality}\\
&=\min\left\{\left\langle \tilde \w_h^k -\tilde \w_h^\ast, \phib^{\tilde\pi_k(s,h)}(s)\right\rangle+\left\langle \tilde \w_h^\ast , \phib^{\tilde\pi_k(s,h)}(s)\right\rangle+\kappa_h(s)\beta\norm{\phib^{\tilde\pi_k(s,h)}(s)}_{\left(\A_h^k\right)^{-1}},H\right\}\label{eq:firsttrandomized}.
\end{align}

Conditioned on event $\Ec_3$ in \eqref{eq:event3} and by the induction assumption, we have
\begin{align}
   -\beta\norm{\phib^{\tilde\pi_k(s,h)}(s)}_{\left(\A_h^k\right)^{-1}}&\leq \left\langle \tilde \w_h^k-\tilde \w_h^\ast , \phib^{\tilde\pi_k(s,h)}(s)\right\rangle+\mathbb{E}_{a\sim\tilde\pi_k(s,h)}\left[\mathbb{E}_{s'\sim \Pb(.|s,a)}\left[\tilde V_{h+1}^k(s')-\tilde V_{h+1}^\ast(s')\right]\right]\nn\\
   &\leq \left\langle \tilde \w_h^k-\tilde \w_h^\ast , \phib^{\tilde\pi_k(s,h)}(s)\right\rangle\label{eq:seconddrandomized}.
\end{align}
By combining \eqref{eq:firsttrandomized} and \eqref{eq:seconddrandomized}, we conclude that
\begin{align}
  \tilde V_h^k(s) &\geq\min\left\{\left\langle \tilde \w_h^\ast , \phib^{\tilde\pi_k(s,h)}(s)\right\rangle+(\kappa_h(s)-1)\beta\norm{\phib^{\tilde\pi_k(s,h)}(s)}_{\left(\A_h^k\right)^{-1}},H\right\}\nn\\
  &=\min\left\{\alpha_h^k(s)\left\langle \tilde \w_h^\ast , \phib^{\pi_\ast(s,h)}(s)\right\rangle+(1-\alpha_h^k(s))\left\langle \tilde \w_h^\ast , \phib^{\pi_0(s,h)}(s)\right\rangle+(\kappa_h(s)-1)\beta\norm{\phib^{\tilde\pi_k(s,h)}(s)}_{\left(\A_h^k\right)^{-1}},H\right\}\nn\\
  &\geq\min\left\{\alpha_h^k(s)\left\langle \tilde \w_h^\ast , \phib^{\pi_\ast(s,h)}(s)\right\rangle+(\kappa_h(s)-1)\beta\norm{\Phi_0^\bot\left(s,\phib^{\tilde\pi_k(s,h)}(s)\right)}_{\left(\A_{h,s}^k\right)^{-1}},H\right\}\nn\\
  &=\min\left\{ \alpha_h^k(s)\left(\left\langle \tilde \w_h^\ast , \phib^{\pi_\ast(s,h)}(s)\right\rangle+(\kappa_h(s)-1)\beta\norm{\Phi_0^\bot\left(s,\phib^{\pi_\ast(s,h)}(s)\right)}_{\left(\A_{h,s}^k\right)^{-1}}\right),H\right\}\label{eq:maxQgeqrandomized}
\end{align}
where the third inequality holds because $\norm{\phib^{\tilde\pi_k(s,h)}(s)}_{\left(\A_h^{k}\right)^{-1}}\geq \norm{\Phi_0^\bot\left(s,\phib^{\tilde\pi_k(s,h)}(s)\right)}_{\left(\A_{h,s}^k\right)^{-1}}$ (see Lemma 3 in \cite{pacchiano2020stochastic} for a proof) and $(1-\alpha_h^k(s))\left\langle\tilde \w_h^\ast,\phib\left(s,a_0\left(s\right)\right)\right\rangle\geq 0$ as the reward is always positive, i.e., $r_h(s,a)\in[0,1]$ for all $(s,a,h)\in\Sc\times\Ac\times[H]$. The second equality follows from the fact that
\begin{align}
    \Phi_0^\bot\left(s,\phib^{\tilde\pi_k(s,h)}(s)\right) =  \alpha_h^k(s)\Phi_0^\bot\left(s,\phib^{\pi_\ast(s,h)}(s)\right)+(1-\alpha_h^k(s))\Phi_0^\bot\left(s,\phib^{\pi_0(s,h)}(s)\right) = \alpha_h^k(s)\Phi_0^\bot\left(s,\phib^{\pi_\ast(s,h)}(s)\right).
\end{align}

Now, we show that $\alpha_{h}^k(s)\geq \frac{\tau-\tau_h(s)}{\tau-\tau_h(s)+2\beta\norm{\Phi_0^\bot\left(s,\phib^{\pi_\ast(s,h)}(s)\right)}_{\left(\A_{h,s}^k\right)^{-1}}}$, which eventually leads to a proper value for $\kappa_h(s)>1$ that guarantees for all $(s,h,k)\in\Sc\times[H]\times[K]$ it holds that $\tilde V_h^\ast(s)\leq \tilde V_h^k(s)$ conditioned on $\tilde\Ec=\Ec_1\cap\Ec_3$. Definitions of $\alpha_h^k(s)$ in \eqref{eq:alpharandomized} and the estimated safe set $\Gamma_h^k(s)$ in \eqref{eq:estimatedsafesetrandomized} imply that for all $(s,h,k)\in\Sc\times[H]\times[K]$, we have

\begin{align}\label{eq:somethingelse}
   \frac{(1-\alpha_h^k(s))\left\langle\phib\left(s,a_0\left(s\right)\right),\tilde\phib\left(s,a_0\left(s\right)\right)\right\rangle}{\norm{\phib\left(s,a_0\left(s\right)\right)}_2}&\tau_h(s)+\alpha_h^k(s)\left[\frac{\left\langle \Phi_0\left(s,\phib^{\pi_\ast(s,h)}(s)\right),\tilde\phib\left(s,a_0\left(s\right)\right)\right\rangle}{\norm{\phib\left(s,a_0\left(s\right)\right)}_2} \tau_h(s)\right.\nn\\
   &+\left.\left\langle \gammab_{h,s}^k , \Phi_0^\bot\left(s,\phib^{\pi_\ast(s,h)}(s)\right)\right\rangle+\beta\norm{\Phi_0^\bot\left(s,\phib^{\pi_\ast(s,h)}(s)\right)}_{\left(\A_{h,s}^k\right)^{-1}}\right] = \tau.
\end{align}

Let $M = \frac{\left\langle \Phi_0\left(s,\phib^{\pi_\ast(s,h)}(s)\right),\tilde\phib\left(s,a_0\left(s\right)\right)\right\rangle}{\norm{\phib\left(s,a_0\left(s\right)\right)}_2} \tau_h(s)+\left\langle \gammab_{h,s}^k , \Phi_0^\bot\left(s,\phib^{\pi_\ast(s,h)}(s)\right)\right\rangle+\beta\norm{\Phi_0^\bot\left(s,\phib^{\pi_\ast(s,h)}(s)\right)}_{\left(\A_{h,s}^k\right)^{-1}}$. Note that due to \eqref{eq:mgreaterthancrandomized}, $M>\tau$, and recall that $\tilde\phib\left(s,a_0\left(s\right)\right) = \frac{\phib\left(s,a_0\left(s\right)\right)}{\norm{\phib\left(s,a_0\left(s\right)\right)}_2}$. Thus, \eqref{eq:somethingelse} gives
\begin{align}
  0<\alpha_{h}^k(s) = \frac{\tau-\tau_h(s)}{M-\tau_h(s)}<1.  
\end{align}

In order to lower bound $\alpha_{h}^k(s)$ (upper bound $M$), we first rewrite $M$ as
\begin{align}
    M = &\frac{\left\langle \Phi_0\left(s,\phib^{\pi_\ast(s,h)}(s)\right),\tilde\phib\left(s,a_0\left(s\right)\right)\right\rangle}{\norm{\phib\left(s,a_0\left(s\right)\right)}_2} \tau_h(s)+\left\langle  \gammab_{h}^{\ast}, \Phi_0^\bot\left(s,\phib^{\pi_\ast(s,h)}(s)\right)\right\rangle+\left\langle \gammab_{h,s}^k -\gammab_{h}^{\ast}, \Phi_0^\bot\left(s,\phib^{\pi_\ast(s,h)}(s)\right)\right\rangle\nn\\
    &+\beta\norm{\Phi_0^\bot\left(s,\phib^{\pi_\ast(s,h)}(s)\right)}_{\left(\A_{h,s}^k\right)^{-1}}\label{eq:Mrandomized},
\end{align}
and show that 
\begin{enumerate}[(a)] 
\item $\frac{\left\langle \Phi_0\left(s,\phib^{\pi_\ast(s,h)}(s)\right),\tilde\phib\left(s,a_0\left(s\right)\right)\right\rangle}{\norm{\phib\left(s,a_0\left(s\right)\right)}_2} \tau_h(s)+\left\langle  \gammab_{h}^{\ast}, \Phi_0^\bot\left(s,\phib^{\pi_\ast(s,h)}(s)\right)\right\rangle\leq \tau$ because
\begin{align}
\hspace{-30pt}\frac{\left\langle \Phi_0\left(s,\phib^{\pi_\ast(s,h)}(s)\right),\tilde\phib\left(s,a_0\left(s\right)\right)\right\rangle}{\norm{\phib\left(s,a_0\left(s\right)\right)}_2} \tau_h(s)+\left\langle  \gammab_{h}^{\ast}, \Phi_0^\bot\left(s,\phib^{\pi_\ast(s,h)}(s)\right)\right\rangle&=\left\langle\gammab_h^\ast,\left\langle  \Phi_0\left(s,\phib^{\pi_\ast(s,h)}(s)\right),\tilde\phib\left(s,a_0\left(s\right)\right) \right\rangle \tilde\phib\left(s,a_0\left(s\right)\right) \right\rangle\nn\\
&+\left\langle\gammab_h^\ast,\Phi_0^\bot\left(s,\phib^{\pi_\ast(s,h)}(s)\right)\right\rangle\nn\\
&=\left\langle\gammab_h^\ast,\Phi_0\left(s,\phib^{\pi_\ast(s,h)}(s)\right)\right\rangle+\left\langle\gammab_h^\ast,\Phi_0^\bot\left(s,\phib^{\pi_\ast(s,h)}(s)\right)\right\rangle\nn\\
  &= \left\langle\gammab_h^\ast,\phib^{\pi_\ast(s,h)}(s)\right\rangle\nn\\
  &\leq \tau. \label{eq:leqtaurandomized}
\end{align}
\item $\left\langle \gammab_{h,s}^k -\gammab_{h}^{\ast}, \Phi_0^\bot\left(s,\phib^{\pi_\ast(s,h)}(s)\right)\right\rangle\leq \beta\norm{\Phi_0^\bot\left(s,\phib^{\pi_\ast(s,h)}(s)\right)}_{\left(\A_{h,s}^k\right)^{-1}}$, because conditioned on $\Ec_1$ in \eqref{eq:event11}, we have
\begin{align}
 \hspace{-30pt}   \left\langle \gammab_{h,s}^k -\gammab_{h}^{\ast}, \Phi_0^\bot\left(s,\phib^{\pi_\ast(s,h)}(s)\right)\right\rangle = \left\langle \gammab_{h,s}^k -\Phi_0^\bot\left(s,\gammab_h^\ast\right), \Phi_0^\bot\left(s,\phib^{\pi_\ast(s,h)}(s)\right)\right\rangle\leq \beta\norm{\Phi_0^\bot\left(s,\phib^{\pi_\ast(s,h)}(s)\right)}_{\left(\A_{h,s}^k\right)^{-1}}.\label{eq:leqbetarandomized}
\end{align}
\end{enumerate}
Now, we combine \eqref{eq:Mrandomized}, \eqref{eq:leqtaurandomized} and \eqref{eq:leqbetarandomized} to conclude that
\begin{align}
    M \leq \tau+2\beta\norm{\Phi_0^\bot\left(s,\phib^{\pi_\ast(s,h)}(s)\right)}_{\left(\A_{h,s}^k\right)^{-1}} \quad\Rightarrow\quad  \alpha_h^k(s)\geq \frac{\tau-\tau_h(s)}{\tau-\tau_h(s)+2\beta\norm{\Phi_0^\bot\left(s,\phib^{\pi_\ast(s,h)}(s)\right)}_{\left(\A_{h,s}^k\right)^{-1}}}.
\end{align}
This lower bound on $\alpha_h^k(s)$ combined with \eqref{eq:maxQgeqrandomized} gives
\begin{align}
    \tilde V_h^k(s)&\geq\min\left\{ \frac{\left(\tau-\tau_h(s)\right)\left(\tilde V_h^\ast(s) +(\kappa_h(s)-1)\beta\norm{\Phi_0^\bot\left(s,\phib^{\pi_\ast(s,h)}(s)\right)}_{\left(\A_{h,s}^k\right)^{-1}}\right)}{\tau-\tau_h(s)+2\beta\norm{\Phi_0^\bot\left(s,\phib^{\pi_\ast(s,h)}(s)\right)}_{\left(\A_{h,s}^k\right)^{-1}}},H\right\}
\end{align}
Let $M_1=\beta\norm{\Phi_0^\bot\left(s,\phib^{\pi_\ast(s,h)}(s)\right)}_{\left(\A_{h,s}^k\right)^{-1}}$. Thus, $\tilde V_h^k(s)\geq \tilde V_h^\ast(s) = \min\left\{ V_h^\ast(s),H\right\}$, if and only if
\begin{align}
 \left(\tau-\tau_h(s)\right)\left( V_h^\ast(s)+(\kappa_h(s)-1)M_1\right)\geq \left(\tau-\tau_h(s)+2M_1\right) V_h^\ast(s),
 \end{align}
 which is true if and only if
 \begin{align}\left(\tau-\tau_h(s)\right)(\kappa_h(s)-1)\geq2 V_h^\ast(s)
&\iff \left(\tau-\tau_h(s)\right)(\kappa_h(s)-1)\geq2H
&\iff \kappa_h(s)\geq\frac{2H}{\tau-\tau_h(s)}+1,
\end{align}
as desired.

\subsection{Proof of Theorem \ref{thm:randomizedSUCBregret}}\label{sec:proofofmaintheoremrandomized}

Conditioned on event $\Ec_3$, for any $(a,s,h,k)\in\Ac\times\Sc\times[H]\times[K]$, we have
\begin{align}
    \tilde Q_h^k(s,a)-\tilde Q_h^{\pi_k}(s,a) &= \left\langle \tilde\w_h^k , \phib(s,a)\right\rangle+\kappa_h(s)\beta\norm{\phib(s,a)}_{\left(\A_h^{k}\right)^{-1}}-\tilde Q_h^{\pi_k}(s,a)\nn\\
    &\leq \mathbb{E}_{s'\sim \Pb(.|s,a)}\left[\tilde V_{h+1}^k(s')-\tilde V_{h+1}^{\pi_k}(s')\right]+\left(1+\kappa_h(s)\right)\beta\norm{\phib(s,a)}_{\left(\A_h^{k}\right)^{-1}}\label{eq:Q-Qrandomized}.
\end{align}
Let $\delta_h^k:=\tilde V_{h}^k(s_h^k)-\tilde V_{h}^{\pi_k}(s_h^k)$ and $\zeta_{h+1}^k:=\mathbb{E}_{a\sim \pi_k(s_h^k,h)}\left[\mathbb{E}_{s'\sim \Pb(.|s_h^k,a)}\left[\tilde V_{h+1}^k(s')-\tilde V_{h+1}^{\pi_k}(s')\right]\right]-\delta_{h+1}^k$. We can write
\begin{align}
    \delta_h^k&=\tilde V_{h}^k(s_h^k)-\tilde V_{h}^{\pi_k}(s_h^k)\nn\\
    &= \min\left\{\max_{\theta\in \Gamma^k_h(s_h^k)}\mathbb{E}_{a\sim\theta}\left[\tilde Q_{h}^k(s_h^k,a)\right],H\right\}-\mathbb{E}_{a\sim\pi_k(s_h^k,h)}\left[\tilde Q_h^{\pi_k}(s_h^k,a)\right]\nn\\
    &\leq \max_{\theta\in \Gamma^k_h(s_h^k)}\mathbb{E}_{a\sim\theta}\left[\tilde Q_{h}^k(s_h^k,a)\right]-\mathbb{E}_{a\sim\pi_k(s_h^k,h)}\left[\tilde Q_h^{\pi_k}(s_h^k,a)\right]\nn\\
    &= \mathbb{E}_{a\sim\pi_k(s_h^k,h)}\left[\tilde Q_{h}^k(s_h^k,a)\right]-\mathbb{E}_{a\sim\pi_k(s_h^k,h)}\left[\tilde Q_h^{\pi_k}(s_h^k,a)\right]\nn\\
    &\leq \mathbb{E}_{a\sim \pi_k(s_h^k,h)}\left[\mathbb{E}_{s'\sim \Pb(.|s_h^k,a)}\left[\tilde V_{h+1}^k(s')-\tilde V_{h+1}^{\pi_k}(s')\right]\right]+\left(1+\kappa_h(s)\right)\beta\mathbb{E}_{a\sim\pi_k(s_h^k,h)}\left[\norm{\phib(s_h^k,a)}_{\left(\A_h^{k}\right)^{-1}}\right] \tag{Eqn.~\eqref{eq:Q-Qrandomized}}\\
    &=\delta_{h+1}^k+\zeta_{h+1}^k+\left(1+\kappa_h(s)\right)\beta \mathbb{E}_{a\sim\pi_k(s_h^k,h)}\left[\norm{\phib(s_h^k,a)}_{\left(\A_h^{k}\right)^{-1}}\right],
\end{align}
Now, conditioning on event $\tilde\Ec$ defined in Theorem \ref{thm:tildeevents}, we bound the cumulative regret as follows:
\begin{align}
    R_K&=\sum_{k=1}^K \tilde V_1^{\ast}(s_1^k)-\tilde V_1^{\pi_k}(s_1^k)\leq \sum_{k=1}^K\delta_1^k\tag{Lemma \ref{lemm:optimismrandomized}}\\
    &\leq \sum_{k=1}^K\sum_{h=1}^H \zeta_{h}^k+\sum_{k=1}^K\sum_{h=1}^H\left(1+\kappa_h(s)\right)\beta\mathbb{E}_{a\sim\pi_k(s_h^k,h)}\left[\norm{\phib(s_h^k,a)}_{\left(\A_h^{k}\right)^{-1}}\right]\nn\\
    &\leq \sum_{k=1}^K\sum_{h=1}^H \zeta_{h}^k+\left(1+\kappa\right)\beta\sum_{k=1}^K\sum_{h=1}^H\mathbb{E}_{a\sim\pi_k(s_h^k,h)}\left[\norm{\phib(s_h^k,a)}_{\left(\A_h^{k}\right)^{-1}}\right].\label{eq:randomizedlaststepregret}
\end{align}
We observe that $\{\zeta_h^k\}$ is a martingale
difference sequence satisfying $|\zeta_h^k|\leq 2H$. Thus, thanks to Azuma-Hoeffding inequality, we have
\begin{align}
    \mathbb{P}\left(\sum_{k=1}^K\sum_{h=1}^H\zeta_{h}^k\leq 2H\sqrt{T\log(dT/\delta)}\right)\geq 1-\delta.\label{eq:randomizedazoma}
\end{align}
In order to bound $\sum_{k=1}^K\sum_{h=1}^H\mathbb{E}_{a\sim\pi_k(s_h^k,h)}\left[\norm{\phib(s_h^k,a)}_{\left(\A_h^{k}\right)^{-1}}\right]$, we define the martingle difference sequence $\iota_h^k:=\mathbb{E}_{a\sim\pi_k(s_h^k,h)}\left[\norm{\phib(s_h^k,a)}_{\left(\A_h^{k}\right)^{-1}}\right] - \norm{\phib(s_h^k,a_h^k)}_{\left(\A_h^{k}\right)^{-1}}$, and note that for any $(h,k)\in[H]\times[k]$, we have $|\iota_h^k|\leq 2/\sqrt{\lambda}$. Thus, thanks to Azuma-Hoeffding inequality, we have
\begin{align}
    \mathbb{P}\left(\sum_{k=1}^K\sum_{h=1}^H\iota_{h}^k\leq 2\sqrt{T\log(dT/\delta)/\lambda}\right)\geq 1-\delta.\label{eq:randomizedazoma2}
\end{align}
Now, we are ready to bound $\sum_{k=1}^K\sum_{h=1}^H\mathbb{E}_{a\sim\pi_k(s_h^k,h)}\left[\norm{\phib(s_h^k,a)}_{\left(\A_h^{k}\right)^{-1}}\right]$ as follows:

\begin{align}
    \sum_{k=1}^K\sum_{h=1}^H\mathbb{E}_{a\sim\pi_k(s_h^k,h)}\left[\norm{\phib(s_h^k,a)}_{\left(\A_h^{k}\right)^{-1}}\right]&\leq 2\sqrt{T\log(dT/\delta)/\lambda} + \sum_{k=1}^K\sum_{h=1}^H \norm{\phib(s_h^k,a_h^k)}_{\left(\A_h^{k}\right)^{-1}}.
\end{align}
In order to bound the second term, we have
\begin{align}
    \sum_{k=1}^K\norm{\phib(s_h^k,a_h^k)}_{\left(\A_h^{k}\right)^{-1}}&\leq \sqrt{K\sum_{k=1}^K\norm{\phib(s_h^k,a_h^k)}^2_{\left(\A_h^{k}\right)^{-1}}}\tag{Cauchy-Schwartz inequality}\\
    &\leq\sqrt{2K\log\left(\frac{\det\left(\A_h^K\right)}{\det\left(\A_h^1\right)}\right)}\label{eq:randomizedfirstine}\\
    &\leq\sqrt{2dK\log\left(1+\frac{K}{d\la}\right)}.\label{eq:randomizedsecondine}
\end{align}
In inequality \eqref{eq:randomizedfirstine}, we used the standard argument in regret analysis of linear bandits \cite{abbasi2011improved} (Lemma~11) as follows: 
\begin{align}\label{eq:randomizedstandardarg}
    \sum_{t=1}^n \min\left(\norm{\y_t}^2_{\Vb_{t}^{-1}},1\right)\leq 2\log\frac{\det \Vb_{n+1}}{\det\Vb_{1}}\quad \text{where}\quad \Vb_n = \Vb_{1}+\sum_{t=1}^{n-1}\y_t\y^\top_t.
\end{align}
In inequality \eqref{eq:randomizedsecondine}, we used Assumption \ref{assum:boundedness} and the fact that $\det(\A)=\prod_{i=1}^d \lambda_i(\A) \leq ({\rm trace}(\A)/d)^d$.

Combining \eqref{eq:randomizedlaststepregret}, \eqref{eq:randomizedazoma}, \eqref{eq:randomizedazoma2}, and \eqref{eq:randomizedsecondine}, we have with probability at least $1-3\delta$
\begin{align}
    R_K\leq 2H\sqrt{T\log(dT/\delta)}+2(1+\kappa)\beta\sqrt{2dHT\log\left(1+\frac{Td}{\delta}\right)/\lambda}.
\end{align}
\section{Finite star convex sets and tractability of the experiments}\label{sec:tractablety}

In this section, we show that if for all $s\in\Sc$, the sets $\Dc(s)=\{\phib(s,a):a\in\Ac\}$ are star convex and \emph{finite} around $\phib\left(s,a_0\left(s\right)\right)$ (see Definition \ref{def:finitestarconvex}), then the optimization problem in Line \ref{line:actionselection} of \SUCB~can be solved efficiently. Thanks to Definition \ref{def:finitestarconvex}, for each $s\in\Sc$, there exist finite number $N$ of vectors $\phib\left(s,a_i\left(s\right)\right)$ such that we can write $\Dc(s_h^k)$ as: $\Dc(s):=\cup_{i=1}^N\left[\phib\left(s,a_0\left(s\right)\right),\phib\left(s,a_i\left(s\right)\right)\right]$, where $\left[\phib\left(s,a_0\left(s\right)\right),\phib\left(s,a_i\left(s\right)\right)\right]$ is the line connecting $\phib\left(s,a_0\left(s\right)\right)$ to $\phib\left(s,a_i\left(s\right)\right)$. Since $\phib\left(s,a_0\left(s\right)\right)\in\Dc_h^k(s):=\left\{\phib(s,a):a\in\Ahk(s)\right\}$, the set $\Dc_h^k(s)$ is also a finite star convex set around $\phib\left(s,a_0\left(s\right)\right)$, and
can be written as $\Dc_h^k(s):=\cup_{i=1}^N\left[\phib\left(s,a_0\left(s\right)\right),\phib\left(s,a_{i,h}^k\left(s\right)\right)\right]$, where $\phib\left(s,a_{i,h}^k\left(s\right)\right)=\alpha^{s,k}_{i,h}\phib\left(s,a_i\left(s\right)\right)+(1-\alpha^{s,k}_{i,h})\phib\left(s,a_0\left(s\right)\right)$ and $\alpha^{s,k}_{i,h}=\max \left\{\alpha\in[0,1]: \alpha\phib\left(s,a_i\left(s\right)\right)+(1-\alpha)\phib\left(s,a_0\left(s\right)\right)\in\Dc_h^k(s)\right\}$, which can be solved by doing line search. The optimization problem at Line \ref{line:actionselection} of Algorithm \ref{alg:SUCB} is equivalent to
\begin{align}
    \max_{\x\in \Dc_h^k(s_h^k)}\left\langle \w_h^k , \x\right\rangle+\kappa_h(s_h^k)\beta\norm{\x}_{\left(\A_h^{k}\right)^{-1}},
\end{align}
which can be executed by optimizing over each line $\left[\phib\left(s_h^k,a_0\left(s_h^k\right)\right),\phib\left(s_h^k,a_{i,h}^k\left(s_h^k\right)\right)\right]$ for all $i\in[N]$.
Note that $\left\langle \w_h^k , \x\right\rangle+\kappa_h(s_h^k)\beta\norm{\x}_{\left(\A_h^{k}\right)^{-1}}$ is a convex function in $\x$. Therefore, its maximum over the line $\left[\phib\left(s_h^k,a_0\left(s_h^k\right)\right),\phib\left(s_h^k,a_{i,h}^k\left(s_h^k\right)\right)\right]$ is achieved at either $\phib\left(s_h^k,a_0\left(s_h^k\right)\right)$ or $\phib\left(s_h^k,a_{i,h}^k\left(s_h^k\right)\right)$, which makes the optimization problem at line \ref{line:actionselection} of Algorithm \ref{alg:SUCB} easy and tractable.

\end{document}